\documentclass[aos,preprint]{imsart}

\RequirePackage[OT1]{fontenc}
\RequirePackage{amsthm,amsmath,natbib}
\RequirePackage[colorlinks]{hyperref}
\usepackage{array}
\usepackage{mathrsfs}
\usepackage{amssymb,bbm}
\usepackage{amsfonts}
\usepackage{amsmath,amssymb}
\usepackage{graphicx}
\usepackage{amscd}
\usepackage{color}
\usepackage{mathrsfs}
\usepackage{latexsym}
\usepackage{array,bm}
\usepackage{epstopdf}
\usepackage{enumerate}
\usepackage{subcaption}
\usepackage{soul}

\usepackage[flushleft]{threeparttable}

\usepackage{textcomp}

\usepackage{algorithm}
\usepackage{algorithmic}

\usepackage{booktabs}

\def\boxit#1{\vbox{\hrule\hbox{\vrule\kern6pt
          \vbox{\kern6pt#1\kern6pt}\kern6pt\vrule}\hrule}}

\newcommand{\bbeta}{\bm{\beta}}

\newcommand{\vF}{\bm{F}}
\newcommand{\bx}{\bm{x}}

\newcommand{\btheta}{\bm{\theta}}

\newcommand{\bW}{\bm{W}}

\newcommand{\bvarsigma}{\bm{\varsigma}}

\newcommand{\bvarepsilon}{\bm{\varepsilon}}
\newcommand{\bY}{\bm{Y}}
\newcommand{\bb}{\bm{b}}
\newcommand{\br}{\bm{r}}
\newcommand{\bX}{\bm{X}}
\newcommand{\bA}{\bm{A}}
\newcommand{\bz}{\bm{z}}

\newcommand{\bZ}{\bm{Z}}
\newcommand{\bQ}{\bm{Q}}

\newcommand{\argmin}{\mathop{\mathrm{argmin}}}

\allowdisplaybreaks[1]

\startlocaldefs
\theoremstyle{plain}

\newtheorem{cond}{Condition}

\newcommand{\Var}{\mathrm{Var}}

\newcommand{\Skew}{\mathrm{Skew}}
\newcommand{\Kurt}{\mathrm{Kurt}}

\newcommand{\E}{\mathbb{E}}

\endlocaldefs

\definecolor{gray}{gray}{0.5}

\begin{document}

\begin{frontmatter}
\title{Big portfolio selection by graph-based conditional moments method}

\begin{aug}

 \author[a]{\fnms{Zhoufan} \snm{Zhu}
 \ead[label=e1]{tylerzzf@163.sufe.edu.cn}},
 \author[b]{\fnms{Ningning} \snm{Zhang}
 \ead[label=e2]{yxl18@mails.tsinghua.edu.cn}}
 \and
 \author[b]{\fnms{Ke} \snm{Zhu}\thanksref{t1}
 \ead[label=e3]{mazhuke@hku.hk}}


\thankstext{t1}{Address correspondence to Ke Zhu: Department of Statistics and Actuarial Science, University of Hong Kong, Hong Kong.  E-mail: mazhuke@hku.hk}

 \affiliation[a]{Shanghai University of Finance and Economics}
 \and
 \affiliation[b]{University of Hong Kong}

\end{aug}
%
%
\begin{abstract}
How to do big portfolio selection is very important but challenging for both researchers and practitioners. In this paper, we propose a new
\underline{gra}ph-based \underline{c}onditional mom\underline{e}nts (GRACE) method to do portfolio selection based on thousands of stocks or more.
The GRACE method first learns the conditional quantiles and mean of stock returns via a factor-augmented temporal graph convolutional network, which guides the learning procedure through a factor-hypergraph built by the set of stock-to-stock relations from the domain knowledge as well as
the set of factor-to-stock relations from the asset pricing knowledge.
Next, the GRACE method learns the conditional variance, skewness, and kurtosis of stock returns from the learned conditional quantiles by using the
quantiled conditional moment (QCM) method. The QCM method is a supervised learning procedure to learn these conditional higher-order moments, so it largely overcomes the computational difficulty from the classical high-dimensional GARCH-type methods. Moreover, the QCM method allows
the mis-specification in modeling conditional quantiles to some extent, due to its regression-based nature.
Finally, the GRACE method uses the learned conditional mean, variance, skewness, and kurtosis to construct several performance measures, which are criteria to sort the stocks to proceed the portfolio selection in the well-known 10-decile framework.
An application to NASDAQ and NYSE stock markets shows that the GRACE method performs much better than its competitors, particularly when the
performance measures are comprised of conditional variance, skewness, and kurtosis.
\end{abstract}




\end{frontmatter}
%
%
\newpage

\setcounter{equation}{0}

\section{Introduction}\label{sec:introduction}

In conjunction with the huge growth of stock market capitalization, the number of existing stocks in the financial market is increasing rapidly nowadays, raising a big challenge to researchers and practitioners on how to do portfolio selection based on thousands of stocks or more.
Suppose there are $N$ different stocks with their prices over $T$ timepoints. Let $r_{i,t}$ denote the return of individual stock $i$ at time $t$ with the conditional mean $\mu_{i,t}\equiv\E\left(r_{i,t} \mid \mathcal{F}_{t-1}\right)$, where $i = 1, ..., N$, $t = 1, ..., T$, and $\mathcal{F}_{t}\equiv\sigma\left(r_{i,s} ; i = 1, ..., N,  s \leq t  \right)$ is the available information set up to time $t$. Conventionally, all considered $N$ stocks are sorted into 10 deciles according to their predicted values of $\mu_{i,t}$ (from the smallest to the largest), and then the long-short portfolio is constructed at time $t-1$ by buying the 10\% highest ranking stocks (decile 10) and selling the 10\% lowest (decile 1); see, for example, \cite{Gu2020EmpiricalAP, Gu2021AutoencoderAP}. However, this conventional way using $\mu_{i,t}$ as the performance measure to sort the stocks has two major shortcomings: First, it ignores the impact of conditional variance $h_{i,t}\equiv \Var \left(r_{i,t} \mid \mathcal{F}_{t-1}\right)$, which
is the risk of uncertainty for guiding portfolio selection under the mean-variance criterion \citep{Markowitz1952} or Sharpe ratio criterion \citep{sharpe1994sharpe}; Second, it does not accommodate the observation that rational investors prefer assets with higher skewness and lower kurtosis in the market \citep{scott1980direction,dittmar2002nonlinear}, implying the necessity of considering
the conditional skewness $s_{i,t}\equiv \Skew \left(r_{i,t} \mid \mathcal{F}_{t-1}\right)$  for the asymmetry risk and conditional kurtosis $k_{i,t}\equiv \Kurt \left(r_{i,t} \mid \mathcal{F}_{t-1}\right)$ for the tail risk to proceed the portfolio selection.

To overcome the two shortcomings aforementioned, we aim to sort the stocks for portfolio selection by using the performance measures below:
\begin{align}
\mbox{Mean--variance (MV): }&\mu_{i,t} - \lambda_1 h_{i,t}; \label{eqn_1}\\
\mbox{Mean--variance with skewness and kurtosis (MVSK): }&\mu_{i,t} - \lambda_1 h_{i,t} + \lambda_2 s_{i,t} - \lambda_3 k_{i,t}; \label{eqn_2}\\
\mbox{Sharpe ratio (SR): }&\mu_{i,t}/\sqrt{h_{i,t}}; \label{eqn_3}\\
\mbox{Sharpe ratio with skewness and kurtosis (SRSK): }&\mu_{i,t}/\sqrt{h_{i,t}} + \lambda_2 s_{i,t} - \lambda_3 k_{i,t}, \label{eqn_4}
\end{align}
where $\lambda_i$, $i = 1, 2, 3$, are positive hyperparameters, and they determine how much penalty one needs to pay for the large values of
$h_{i,t}$ and $k_{i,t}$ or how much reward one can gain for the large values of $s_{i,t}$. To implement these four performance measures, we need to learn $h_{i,t}$, $s_{i,t}$, and $k_{i,t}$ dynamically for $N$ stocks. When $N=1$, these three higher-order conditional moments are studied by using some variants of univariate generalized autoregressive conditional heteroskedasticity (GARCH) model  \citep{engle1982autoregressive,bollerslev1986generalized}; see, for example, \cite{Jondeau2003ConditionalVS}, \cite{leon2005autoregresive}, \cite{Len2020ModelingAR}, and references therein. However, those univariate GARCH-type methods have the risk of model mis-specification and the instability of model estimation particularly when the dynamics of $s_{i,t}$ and $k_{i,t}$ are considered.
When $N$ is large (say, e.g., $N=1000$), no clear feasible manner so far has been offered in the literature to
estimate $h_{i,t}$, $s_{i,t}$, and $k_{i,t}$ by using high-dimensional GARCH-type models,
which are formed to study the dynamics of $\br_t\equiv (r_{1,t},...,r_{N,t})'$.
The cause of this dilemma is that the high-dimensional GARCH-type models
have to assume a certain distribution of $\br_{t}$ to estimate $h_{i,t}$, $s_{i,t}$, and $k_{i,t}$, however, the resulting log-likelihood function
is too complex to be optimized. For example, the optimization of the commonly used Gaussian log-likelihood function needs to
invert many $N\times N$-dimensional variance-covariance matrices, and this task becomes computationally infeasible for large $N$ cases.
\cite{pakel2021fitting} propose a composite likelihood estimation (CLE) method for the parsimonious scalar BEKK model to estimate $h_{i,t}$.
Since the CLE method is based on all pairwise Gaussian log-likelihood functions with the order $O(N^2)$, it becomes computationally burdensome when $N$ is in thousands, and meanwhile, it may not adequately capture the dependence among stock returns at the price of pairwise technique.
One way to further reduce the computational burden is to estimate $h_{i,t}$ by the equation-by-equation (EbE) method, as done for the parsimonious scalar DCC model \citep{francq2016estimating,Engle2019LargeDC}. The EbE method is feasible in terms of computation but inefficient in terms of prediction, since it totally ignores the dependence among stock returns.


This paper contributes to the literature by proposing a new \underline{gra}ph-based \underline{c}onditional mom\underline{e}nts (GRACE) method for portfolio selection under four performance measures in (\ref{eqn_1})--(\ref{eqn_4}). The GRACE method has two core engines.
Its first engine is to study the conditional quantiles of $r_{i,t}$ for $i=1,...,N$ and $t=1,...,T$ by a graph-based quantile model,
which can be directly estimated via the quantile loss function \citep{koenker1978regression}.
Our graph-based quantile model is based on a new factor-augmented temporal graph convolutional network (FTGCN),
and thus it is called the FTGCN-based quantile model. This FTGCN-based quantile model uses the stock and factor features to extract both temporal and spatial information for all stocks, and then takes the extracted information to learn the conditional quantiles under the guidance of a factor-augmented hypergraph.
The factor-augmented hypergraph is neither random nor time-variant, and it combines the domain knowledge of the multiple types of relation between any two stocks and the asset pricing knowledge of the impact of common factors on all stocks.
Our factor-augmented hypergraph has a linkage with the hypergraph in TGCN (\citealp{Feng2019TemporalRR}) that also exploits the domain knowledge to
build the graph structure among stocks, where the domain knowledge comes from
the public information of the stocks (e.g., the industrial background, financial statement, and shareholder information),
and its usefulness has been well documented by \cite{Livingston1977INDUSTRYMO}, \cite{Cohen2008economic}, \cite{Lee2019Technological}, \cite{Aaron2021Where}, and many others.
However, the hypergraph in TGCN overlooks an important fact from the asset pricing literature that
some common factors can globally affect all stocks in the market (\citealp{Fama1993CommonRF,fama2015five,Fama2018Choosing,griffin2002fama,hou2011factors}).
This asset pricing knowledge is obviously as informative as the domain knowledge, and it motivates the
design of our factor-augmented hypergraph.
Using the similar idea above, our GRACE method further proposes a FTGCN-based mean model to estimate $\mu_{i,t}$.

Based on the estimated conditional quantiles of $r_{i,t}$ at $K$ different quantile levels from our FTGCN-based quantile model, the second engine of our GRACE method is to estimate $h_{i,t}$, $s_{i,t}$, and $k_{i,t}$ via their corresponding quantiled conditional moments (QCMs) in \cite{QCM}. The QCM method estimates $h_{i,t}$, $s_{i,t}$, and $k_{i,t}$ through the ordinary least squares (OLS) estimator of a linear regression model, which is constructed by those estimated conditional quantiles.
The formulation of this linear regression model stems naturally from
the Cornish-Fisher expansion \citep{Fisher1938148MA}, which exhibits a fundamental relationship between conditional quantiles and conditional moments. In principle, the QCM method transforms the estimation of $h_{i,t}$, $s_{i,t}$, and $k_{i,t}$ to that of
conditional quantiles, and this brings us two substantial advantages over the GARCH-type method.
First, the QCM method is easy-to-implement as long as the estimated conditional quantiles of $r_{i,t}$ are provided.
Note that our FTGCN-based quantile model can estimate conditional quantiles of $r_{i,t}$ for large $N$ and $T$ cases by
a supervised learning through the use of quantile loss function. Therefore, unlike the estimation of high-dimensional GARCH-type models,
no assumption on the distribution of $\br_{t}$ is needed to estimate our FTGCN-based quantile model. This is the reason why the QCM method can
make the estimation of higher-order moments feasible for large $N$ cases,
although it needs to estimate the quantile model $K$ different times.
Second, the QCM method largely alleviates the risk of model mis-specification, since the QCMs of
$h_{i,t}$, $s_{i,t}$, and $k_{i,t}$ are proposed without any estimator of $\mu_{i,t}$, and more importantly, they are
consistent even when the conditional quantile estimators of $r_{i,t}$ are biased to some extent.
In this sense, our FTGCN-based quantile model could generate consistent QCMs,
as long as its specification does not largely deviate from the true specification of conditional quantile of $r_{i,t}$.


We apply our GRACE method to construct long-short portfolios based on 1026 and 1737 stocks in NASDAQ and NYSE, respectively.
To build the factor-augmented hypergraph, we use the Wiki company-based relations \citep{Feng2019TemporalRR} as the domain knowledge to specify the multiple types of relation between any two stocks, and at the same time, we take the Fama-French five factors \citep{fama2015five} as the asset pricing knowledge to capture the impact of common factors on all stocks. From an economic viewpoint, our empirical results are encouraging in four aspects. First, all of the MV, MVSK, SR, and SRSK portfolios have larger values of out-of-sample annualized SR than the M portfolio in the GRACE method.
Second, the SRSK portfolio from the GRACE method performs the best, and its values of out-of-sample annualized SR
are $4.81$ and $3.48$ in NASDAQ and NYSE, respectively, which are $236\%$ and $21\%$ higher than those of the M portfolio from the benchmark method in \cite{Feng2019TemporalRR}. Third, the GRACE method always dominates the simple GRACE method in portfolio selection by a wide margin, where the
simple GRACE method adopts the linear structure \citep{zhu2017network,zhu2019network} to extract the information from stock and factor features to learn the conditional quantiles and mean of stock returns.
Fourth, regardless of performance measure, the portfolios from the GRACE method have a more robust performance than those from its competitors over the set of stock-to-stock relations, the choice of hyperparameters, and the level of transection cost.
All of these aforementioned findings indicate the importance of using the higher-order conditional moments to form the performance measure, the asset pricing knowledge to build the hypergraph, and the network structure to extract the feature information. From a statistical viewpoint, the conditional moments learned from the GRACE method are largely valid and better than those from other competing methods, shedding light on the advantage of GRACE method in portfolio selection.

The remaining paper is organized as follows. Section \ref{sec:method} presents our entire methodology, including the network architecture of FTGCN, the training procedure of FTGCN-based quantile and mean models, the formal estimation procedure of the QCMs, and the implementation details of the GRACE method. Section \ref{sec:empirical} presents our empirical studies of big portfolio selection in NASDAQ and NYSE stock markets. Concluding remarks are offered in Section \ref{sec:conclusion}.

\section{Methodology}\label{sec:method}


\subsection{Graph-based Learning for Conditional Quantiles} Let $\bQ_t(\tau) = (Q_{1,t}(\tau), ..., Q_{N,t}(\tau))'$ be the high-dimensional vector of  $\tau$-th conditional quantiles, where $Q_{i,t}(\tau)$ is the $\tau$-th conditional quantile of $r_{i,t}$ given $\mathcal{F}_{t-1}$. In this paper, we study $\bQ_t(\tau)$ by a new FTGCN-based quantile model defined as
\begin{align}\label{bQ_t}
    \bQ_t(\tau) = f(\bX_{t-1}; \mathcal{G}, \btheta_\tau),
\end{align}
where $\bX_{t-1} \in \mathcal{R}^{(N + B) \times P \times S}$ is a feature tensor built on $\mathcal{F}_{t-1}$ including the information of $N$ stocks and $B$ factors, and
$f(\cdot; \mathcal{G}, \btheta_\tau): \mathcal{R}^{(N + B) \times P \times S}\to \mathcal{R}^{N\times 1}$ is the FTGCN depending on a factor-augmented hypergraph $\mathcal{G}$ and a vector of unknown parameters $\btheta_\tau$.
Here, $\bX_{t-1}=[\bX_{1,t-1},...,\bX_{N,t-1},$ $\bX_{N+1, t-1}, ..., \bX_{N+B, t-1}]$ with
$\bX_{i,t-1}\in\mathcal{R}^{P\times S}$ having its $s$-th column $\bx_{i,t-1-S+s}\in\mathcal{R}^{P}$,
where $\bX_{i,t-1}$ for $i=1,...,N$ is the feature matrix for stock $i$,
$\bX_{N+b,t-1}$ for $b=1,...,B$ is the feature matrix for factor $b$,
$P$ with a potential high dimension is the number of stock or factor features, $S$ is the number of lagged values of each feature, and $\bx_{i,t}$ (or $\bx_{N+b,t}$) is the vector of $P$ different features of stock $i$ (or factor $b$) at time $t$.
Below, we show the four modules to construct the FTGCN $f(\cdot; \mathcal{G}, \btheta_\tau)$.

\subsubsection{Module I: Feature Extraction} In the first module, we employ a one-layer long short-term memory (LSTM) network \citep{Hochreiter1997LongSM} to extract the temporal embedding $\bx_{i,t}^{L}\in\mathcal{R}^{d}$ from the feature matrix $\bX_{i,t-1}$ at time $t-1$. Specifically, we let
$\underline{\bx}_{i,t-1}^s=\bx_{i,t-1-S+s}$ and compute the hidden state vectors $\bm{h}_{i,t}^{s}$, $s=1,...,S$, recursively from
the LSTM network:
\begin{align}\label{LSTM}
\begin{split}
    \bm{z}_{i,t}^s &= \mathrm{tanh} (\bW_{1x} \underline{\bx}_{i,t-1}^s + \bW_{1h} \bm{h}_{i,t}^{s-1} + \bb_1),\\
    \bm{i}_{i,t}^s &= \mathrm{sigmoid} (\bW_{2x} \underline{\bx}_{i,t-1}^s + \bW_{2h} \bm{h}_{i,t}^{s-1} + \bb_2),\\
    \bm{f}_{i,t}^s &= \mathrm{sigmoid} (\bW_{3x} \underline{\bx}_{i,t-1}^s + \bW_{3h} \bm{h}_{i,t}^{s-1} + \bb_3),\\
    \bm{c}_{i,t}^s &= \bm{f}_{i,t}^s \odot \bm{c}_{i,t}^{s-1} + \bm{i}_{i,t}^s \odot \bm{z}_{i,t}^s,\\
    \bm{o}_{i,t}^s &= \mathrm{sigmoid} (\bW_{4x} \underline{\bx}_{i,t-1}^s + \bW_{4h} \bm{h}_{i,t}^{s-1} + \bb_4),\\
    \bm{h}_{i,t}^{s} &= \bm{o}_{i,t}^s \odot \mathrm{tanh}(\bm{c}_{i,t}^s),
\end{split}
\end{align}
where $\bW_{1x}, \bW_{2x}, \bW_{3x}, \bW_{4x} \in \mathcal{R}^{d \times P}$ and $\bW_{1h}, \bW_{2h}, \bW_{3h}, \bW_{4h} \in \mathcal{R}^{d \times d}$ are matrices of weight parameters, $\bb_1, \bb_2, \bb_3, \bb_4 \in \mathcal{R}^d$ are vectors of bias parameters, $d$ is the number of hidden units controlling the network complexity, $\mathrm{tanh}(\cdot)$ and $\mathrm{sigmoid}(\cdot)$ are two entry-wise vector-valued functions, $\odot$ is element-wise production operation, and the initial values $\bm{c}_{i,t}^0$ and $\bm{h}_{i,t}^0$ are conventionally set to the $d$-dimensional vector of zeros. Then, our temporal embedding $\bx_{i,t}^{L}$ is taken as the last hidden state vector $\bm{h}_{i,t}^{S}$ (the output of LSTM network), that is,
\begin{align}\label{x_lstm}
    \bx_{i,t}^{L} = \bm{h}_{i,t}^{S} = h(\bX_{i,t-1}; \btheta_{L}) \in \mathcal{R}^{d},
\end{align}
where $h(\cdot; \btheta_{L})$ is the LSTM network in (\ref{LSTM}) indexed by $\btheta_{L}$, and $\btheta_{L}$ contains all the parameters in $\{\bW_{jx}, \bW_{jh}, \bb_j: j=1,2,3,4\}$. Clearly, the purpose of this module is to extract an expressive vector $\bx_{i,t}^{L}$, which stores the long-term temporal information of $P$ features up to time $t-1$. It is expected that all the
 temporal information carried by $\{\bx_{1,t}^{L},...,\bx_{N+B,t}^{L}\}$ can help us to predict the behavior of
future returns $\{r_{1,t},...,r_{N,t}\}$ at time $t$.

\subsubsection{Module II: Hypergraph Construction} Besides the temporal information from the stock features, the spatial information (i.e., interdependence relations) among all stocks is also important for predictions. For example, (i) MSFT LLC and Google LLC could have an industry-specific relation, since both of them belong to ``Computer Software: Programming'' industry; and (ii) Boeing Inc. and United Airlines Inc. could have a corporate relation, in view of the fact that Boeing Inc. produces Boeing airplanes for United Airlines Inc. Needless to say, these stock-to-stock (S2S) relations are informative and should not be ignored. In general, we can have $M$ different types of S2S relation (denoted by $\bm{E}_{stock} = \{e_1, ..., e_M\}$) between any two stocks based on the domain knowledge.

Along with the S2S relations, the factor-to-stock (F2S) relations also exist in the market, since the stock returns can move together driven by the common factors; see the vast evidence in the asset pricing literature (\citealp{Fama2018Choosing,lettau2020estimating,Gu2021AutoencoderAP}). The F2S relations convey the spatial information from factors to stocks, and they are highly possible to be factor-specific. Therefore, based on the asset pricing knowledge, we consider $B$ different F2S relations (denoted by $\bm{E}_{factor} = \{e_{M+1}, ..., e_{M+B}\}$), where the F2S relation $e_{M+b}$ is induced by factor $b$.

To describe all of S2S and F2S relations above, we build a factor-augmented hypergraph
\begin{align}\label{G_graph}
\mathcal{G} = (\bm{V}, \bm{A}),
\end{align}
where $\bm{V} = \{\bm{V}_{stock}, \bm{V}_{factor}\}$ is the set of vertices, and $\bm{A} = \{\bm{A}_{stock}, \bm{A}_{factor}\}$ is the set of adjacency matrices. Here, $\bm{V}_{stock} = \{1, ..., N \}$ is the set of stock vertices with the vertex $i \in \bm{V}_{stock}$ representing the stock $i$, $\bm{V}_{factor} = \{N+1, ..., N+B\}$ is the set of factor vertices with the vertex $N+b \in \bm{V}_{factor}$ representing the factor $b$,
 $\bm{A}_{stock}=\{\bm{A}_1,...,\bm{A}_{M}\}$ is the set of adjacency matrices with the matrix $\bA_m\in \bm{A}_{stock}$ representing the S2S relation $e_m$, and $\bm{A}_{factor} = \{\bm{A}_{M + 1}, ..., \bm{A}_{M + B}\}$ is the set of adjacency matrices with the matrix $\bA_{M+b}\in \bm{A}_{factor}$ representing the F2S relation $e_{M+b}$, where
 $\bA_m$ has its $(i,j)$-th entry
\begin{align*}
    a_{i,j,m} =
    \begin{cases}
        1, & \text{if there is an S2S relation } e_m \text{ between vertices } i\in \bm{V}_{stock} \text{ and } j\in \bm{V}_{stock}, \\
        0, & \text{otherwise},
    \end{cases}
\end{align*}
and $\bA_{M+b}$ has its $(i,j)$-th entry
\begin{align*}
    a_{i,j, M+b} =
    \begin{cases}
        1, & \text{if } i \in \bm{V}_{stock} \mbox{ and } j = N+b\in\bm{V}_{factor} \text{ or } j \in \bm{V}_{stock} \mbox{ and }  i = N+b\in\bm{V}_{factor}, \\
        0, & \text{otherwise}.
    \end{cases}
\end{align*}
According to the definitions of $\bA_m$ and $\bA_{M+b}$, the factor-augmented hypergraph $\mathcal{G}$ ensures that (i) two stock vertices in $\bm{V}_{stock}$ are linked when they have up to $M$ different S2S relations; and
(ii) each factor vertex in $\bm{V}_{factor}$ is linked to all of stock vertices in $\bm{V}_{stock}$ indicating the corresponding F2S relation.
Since our main target is to study the dynamics of stocks rather than factors, we assume that
there has no linkage between any two factor vertices in $\bm{V}_{factor}$ for simplicity, and our analysis results below
do not change even when the factor vertices are allowed to have connections.

 In sum, the overall relation between any two vertices $i$ and $j$ in $\mathcal{G}$ can be represented by the vector
\begin{align}\label{vector_a}
    \bm{a}_{i,j} = (a_{i,j,1}, ...,a_{i,j,M},a_{i,j,M+1},..., a_{i,j,M + B})' \in \mathcal{R}^{M + B},
\end{align}
where the first $M$ entries and the remaining $B$ entries carry the information of S2S relations and F2S relations, respectively.


\subsubsection{Module III: Hypergraph Learning} Having known the relations among all stocks and factors in $\mathcal{G}$, it is natural to capture how much temporal information the stock $i$ can receive from its linked stocks and factors. To fulfill this goal, we define the aggregated temporal embedding for stock $i$ as
\begin{align}\label{TGC}
    \bx_{i, t}^{P} = \sum_{j \in \bm{V}_{stock},j \neq i} \frac{g\left(\bm{a}_{i,j}, \bx_{i,t}^{L}, \bx_{j,t}^{L}; \mathcal{G}, \btheta_{P} \right)}{d_{j}} \bx_{j,t}^{L} + \sum_{j \in \bm{V}_{factor}} \frac{g\left(\bm{a}_{i,j}, \bx_{i,t}^{L}, \bx_{j,t}^{L}; \mathcal{G}, \btheta_{P} \right)}{N} \bx_{j,t}^{L}
    \in \mathcal{R}^d,
\end{align}
where $i=1,...,N$, $\bm{a}_{i,j}$ in (\ref{vector_a}) represents the overall relation between vertices $i$ and $j$, $\bx_{i,t}^{L}$ and $\bx_{j,t}^{L}$ are the temporal embeddings extracted by (\ref{x_lstm}), $g(\cdot): \mathcal{R}^{M+B}\times \mathcal{R}^{d}\times \mathcal{R}^{d}\to \mathcal{R}$ is a nonlinear function indexed by $\btheta_{P}$ to measure the intensity of interplay between vertices $i$ and $j$, and $d_j = \sum_{m = 1}^M \sum_{i \neq j} a_{i,j,m}$ is the number of stocks linked with stock $j$. Following \cite{Feng2019TemporalRR}, we take
\begin{align}\label{strength}
   g\left(\bm{a}_{i,j}, \bx_{i,t}^{L}, \bx_{j,t}^{L}; \mathcal{G}, \btheta_{P} \right) = \mathrm{softmax}\left(\bW_5 (\bx_{i,t}^{L\prime}, \bx_{j,t}^{L\prime}, \bm{a}_{i,j}')' + b_5 \right) \in \mathcal{R},
\end{align}
where $\bW_5 \in \mathcal{R}^{1 \times (M+B+2d)}$ is a vector of weight parameters, $b_5\in\mathcal{R}$ is a bias parameter,
 $\btheta_{P}$ contains all the parameters in $\bW_5$ and $b_5$, and
$\mathrm{softmax}(\cdot)$ is used to normalize the value of $g(\cdot)$ into $(0, 1)$. The specification of $g(\cdot)$ in (\ref{strength}) has two merits: First, it allows the intensity of interplay between any two vertices to be stock-, factor-, and relation-specific; Second,
it aims to capture some missing relations that are not described in $\bm{a}_{i,j}$ (i.e., $\bm{a}_{i,j}\equiv 0$)
but presented by the similarity of $\bx_{i,t}^{L}$ and $\bx_{j,t}^{L}$, since the term
$\bW_5 (\bx_{i,t}^{L\prime}, \bx_{j,t}^{L\prime}, \bm{a}_{i,j}')'$ is still informative even
when $\bm{a}_{i,j}\equiv 0$.

As a temporal graph convolution (TGC), the third module combines the temporal embedding $\bx_{i,t}^L$ in (\ref{x_lstm}) and the aggregated temporal embedding $\bx_{i,t}^P$ in (\ref{TGC}) to form
\begin{align}\label{x_tgc}
    \bx_{i,t}^{TGC} = \left(\bx_{i,t}^{L \prime}, \bx_{i,t}^{P\prime} \right)' \in \mathcal{R}^{2d} \mbox{ for }i=1,...,N.
\end{align}
The advantage of using $\bx_{i,t}^{TGC}$ is apparent, since $\bx_{i,t}^{TGC}$ captures the spatial and temporal information of stock features simultaneously.

\subsubsection{Module IV: Quantile Output}
\begin{figure}[!h]
    \centering
    \includegraphics[width = \linewidth]{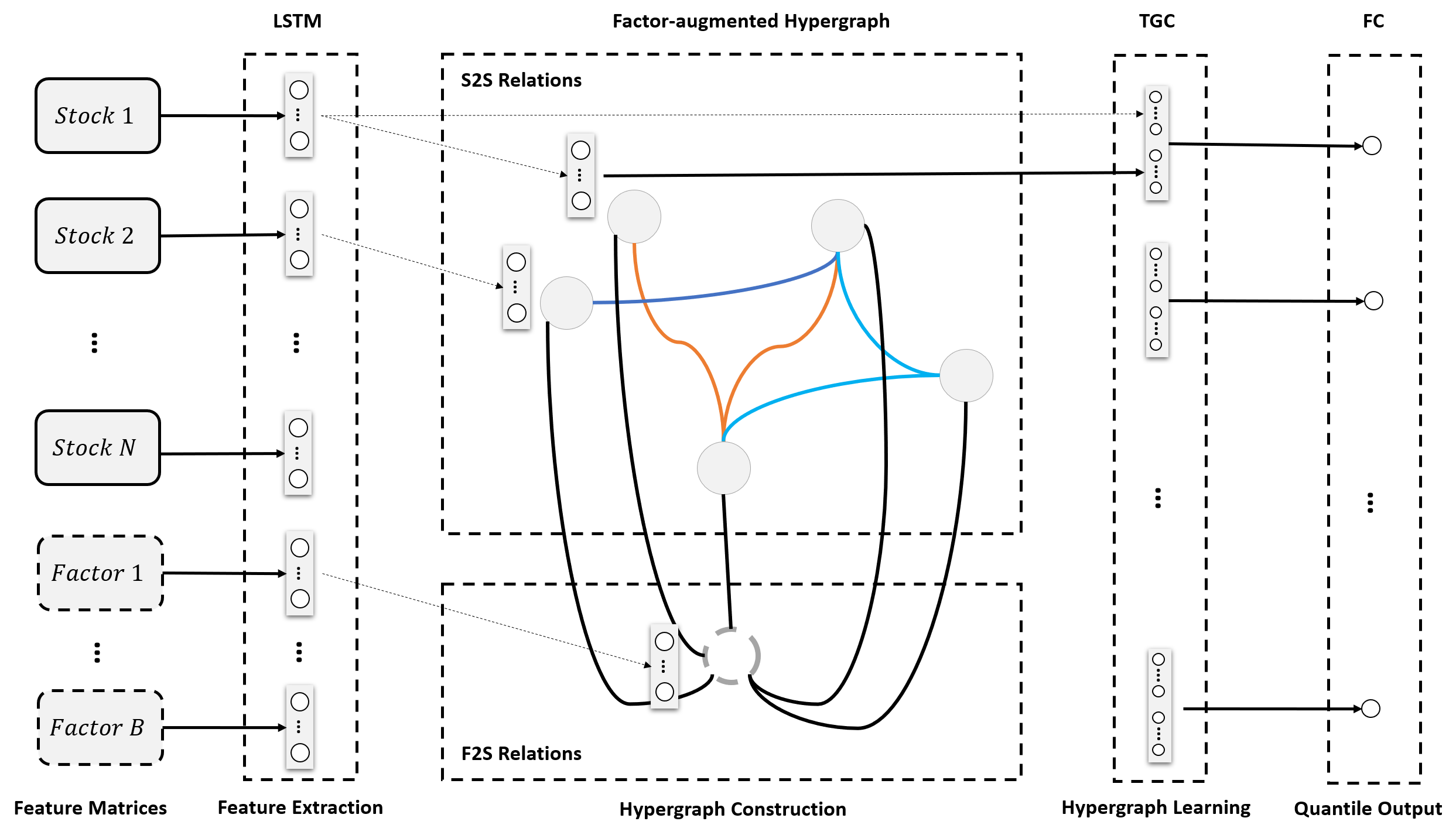}
    \caption{The architecture of FTGCN-based quantile model.}
    \label{fig:network}
\end{figure}

Our last module applies a fully connected (FC) network to revise the spatial-temporal information
$\bx_{i,t}^{TGC}$ in (\ref{x_tgc}) to $\bQ_t(\tau)$ in (\ref{bQ_t}).
Let $\btheta_C$ contain all the parameters in $\bW_6$ and $b_6$, where $\bW_6 \in \mathcal{R}^{1 \times 2d}$ is a vector of weight parameters, and $b_6\in\mathcal{R}$ is a bias parameter. Then, we set
the form of FTGCN as
\begin{align}\label{FTGCN}
\begin{split}
&f(\bX_{t-1}; \mathcal{G}, \btheta)\equiv (f_1(\bX_{t-1}; \mathcal{G}, \btheta),...,f_{N}(\bX_{t-1}; \mathcal{G}, \btheta))'\\
&\mbox{ with }f_{i}(\bX_{t-1}; \mathcal{G}, \btheta)=\bW_6 \bx_{i,t}^{TGC} + b_6 \mbox{ for }i=1,...,N,
\end{split}
\end{align}
where $\mathcal{G}$ is the factor-augmented hypergraph in (\ref{G_graph}), and $\btheta$ contains all the parameters in $\btheta_L$, $\btheta_P$, and $\btheta_C$. Consequently, our FTGCN-based quantile model has the specification:
\begin{align}\label{final_f}
\bQ_t(\tau) = f(\bX_{t-1}; \mathcal{G}, \btheta_\tau)
\mbox{ with }Q_{i,t}(\tau)= f_i(\bX_{t-1}; \mathcal{G}, \btheta_\tau)\mbox{ for }i=1,...,N
\end{align}
(see its network architecture in Fig\,\ref{fig:network}), where $f(\bX_{t-1}; \mathcal{G}, \btheta_\tau)$ and $f_i(\bX_{t-1}; \mathcal{G}, \btheta_\tau)$ are defined as in (\ref{FTGCN}).


\subsubsection{Estimation of the FTGCN-based Quantile Model}

As $Q_{i,t}(\tau)$ is the $\tau$-th conditional quantile of $r_{i,t}$ given $\mathcal{F}_{t-1}$, we estimate $\btheta_{\tau}$ in (\ref{final_f}) by the following quantile estimator:
\begin{align}\label{qr_loss}
    \widehat{\btheta}_\tau &= \argmin_{\btheta_\tau} \frac{1}{NT} \sum_{i = 1}^N \sum_{t = 1}^T \rho_\tau \left(r_{i,t} - f_i(\bX_{t-1}; \mathcal{G}, \btheta_\tau) \right)
     \equiv  \argmin_{\btheta_\tau} \frac{1}{T} \sum_{t = 1}^T \ell(\br_t, \bX_{t-1}; \mathcal{G}, \btheta_\tau, \tau),
\end{align}
where $\rho_\tau(x) = x[\tau - I(x < 0)]$ is the quantile loss function \citep{koenker1978regression} with $I(\cdot)$ being the indicator function.
Due to the massive data volume, we adopt the adaptive moment estimation (Adam) algorithm in \cite{Kingma2015AdamAM} to compute $\widehat{\btheta}_\tau$ in \eqref{qr_loss}; see Algorithm \ref{alg:adam_quantile} for the details.
Using $\widehat{\btheta}_\tau$, we obtain
\begin{align}\label{hat_bQ_t}
    \widehat{\bQ}_{t}(\tau) = f(\bX_{t-1}; \mathcal{G}, \widehat{\btheta}_\tau),
\end{align}
which is the estimator of $\bQ_{t}(\tau)$ at the quantile level $\tau$.

\begin{algorithm}[h!]
\caption{The training procedure of $\widehat{\btheta}_\tau$ by the Adam algorithm.}\label{alg:adam_quantile}
\begin{algorithmic}[1]
\REQUIRE ~~\\
    The sample: $\{(\bm{r}_{t}, \bX_{t-1})\}$;\\
    The initial value of network parameters in the FTGCN-based quantile model: $\btheta_{\tau}^{(0)}$;\\
    The factor-augmented hypergraph: $\mathcal{G}$;\\
    Hyperparameter: learning rate $\gamma$;\\
\STATE $l = 0$;
\REPEAT
\STATE $(\br_l, \bX_{l-1})$ $\leftarrow$ draw a random data point from $\{(\br_t, \bX_{t-1})\}$ (A cross-sectional minibatch);
\STATE $\bm{g}^{(l)}_{\tau} \leftarrow \nabla_{\btheta_\tau} \left[\ell(\br_l, \bX_{l-1}; \mathcal{G}, \btheta^{(l)}_\tau, \tau) \right]$ (Gradients of minibatch estimator);
\STATE $\btheta_\tau^{(l + 1)} \leftarrow$ update parameters using learning rate $\gamma$ and gradients $\bm{g}^{(l)}_{\tau}$ (Adam);
\STATE $l \leftarrow l+1$;
\UNTIL{convergence of parameters $\btheta_\tau^{(l + 1)}$};
\ENSURE ~~\\
    The value of $\btheta_\tau^{(l + 1)}$, which is taken as the quantile estimator $\widehat{\btheta}_\tau$.
\end{algorithmic}
\end{algorithm}

\subsubsection{Comparison with the Existing Models}

Our FTGCN-based quantile model in (\ref{final_f}) has a linkage with the TGCN-based model in \cite{Feng2019TemporalRR} with regard to the network structure. As the pioneering work, the TGCN-based model applies the domain knowledge to construct a hypergraph for taking multiple types of S2S relation into account. The main difference between the FTGCN-based quantile model and the TGCN-based model is two-fold.
First, the FTGCN-based quantile model aims to learn the conditional quantiles of $r_{i,t}$, whereas the TGCN-based model
focuses on the conditional mean of $r_{i,t}$. Second, the FTGCN-based quantile model incorporates the F2S relations to build the factor-augmented hypergraph, but the TGCN-based model does not consider this kind of important information in its hypergraph.

Besides our FTGCN-based quantile model, many other models are existing in the literature to study the conditional quantile of high-dimensional data;
see, for example, \cite{koenker2004quantile}, \cite{kato2012asymptotics}, and \cite{galvao2016smoothed}
for the quantile individual fixed effects models, \cite{ando2020quantile}, \cite{Chen2021QuantileFM}, \cite{ma2021estimation}, and \cite{YangQuantile} for the quantile factor models, and \cite{TENET}, \cite{zhu2019network}, and \cite{xu2022dynamic} for the quantile network models.
However, except for the factor-augmented dynamic network quantile regression (FDNQR) model in \cite{xu2022dynamic}, none of the aforementioned models takes the domain knowledge and asset pricing knowledge simultaneously into account to guide the estimation of conditional quantile.
Specifically,  the FDNQR model uses the domain knowledge to propose a weighted adjacency matrix $\bW\in\mathcal{R}^{N\times N}$ with the $(i,j)$-th entry $w_{i,j}$, where $w_{i,j}=a_{i,j}/n_i$, $n_i=\sum_{j=1}^{N}a_{i,j}$, $a_{i,j}=1$ if the stock $i$ has the connection with another stock $j$, and $a_{i,j}=0$ otherwise. Based on $\bW$, the FDNQR model assumes
\begin{align}\label{NQAR}
    Q_{i,t}(\tau) = \alpha_{\tau} + \bbeta_{\tau}'\bz_{i} + \gamma_{\tau} \sum_{j=1}^{N} w_{i,j} r_{j,t-1} +
    \zeta_{\tau} r_{i, t-1}+\sum_{s=1}^{S}\bvarsigma_{s,\tau}'\vF_{t-s},
\end{align}
where $\alpha_{\tau}\in\mathcal{R}$, $\bbeta_{\tau}\in\mathcal{R}^{Q\times 1}$,
$\gamma_{\tau}\in\mathcal{R}$, and $\bvarsigma_{s,\tau}\in\mathcal{R}^{B\times 1}$ are quantile regression coefficients,
$\bz_{i}\in\mathcal{R}^{Q\times 1}$ is a $Q$-dimensional vector of time-invariant stock features, and
$\vF_t=(f_{1,t},...,f_{B,t})'\in\mathcal{R}^{B\times 1}$ is a $B$-dimensional vector of time-variant factors.
In model \eqref{NQAR}, $\alpha_{\tau}$ is the constant intercept term for all stocks, $\bbeta_{\tau}$ is the constant intensity of
the impact from stock features on stock $i$, $\gamma_{\tau}$ is the constant intensity of spatial impact on stock $i$ caused by its connected stocks, $\zeta_{\tau}$ is the constant intensity of temporal impact on stock $i$ caused by its lagged term, and $\bvarsigma_{s,\tau}$
is the constant intensity of factor impact on all stocks caused by the lagged factors. Clearly, our FTGCN-based quantile model is much more general than model (\ref{NQAR}), since it captures multiple types of relation separately, extracts the information of time-variant
stock and factor features in a non-linear way, and allows for the time-variant heterogenous intensity of spatial and temporal impacts on each stock caused by either its connected stocks or factors.

Note that model \eqref{NQAR} does not include the contemporaneous variables in the original FDNQR model for the purpose of prediction, and it nests the network quantile autoregressive model in \cite{zhu2019network}. As the contemporaneous variables are absent, model \eqref{NQAR} now can be consistently estimated by using the quantile loss function as in \cite{zhu2019network}.

\subsection{Graph-based Learning for Conditional Mean}\label{sec_mean}

So far, we have introduced the FTGCN to learn the conditional quantile of $r_{i,t}$.
Following the similar idea, we can learn the conditional mean of $r_{i,t}$ by an FTGCN-based mean model:
\begin{align}\label{mean_model}
    r_{i,t}=f(\bX_{t-1}; \mathcal{G}, \btheta_\mu)+\varepsilon_{i,t},
\end{align}
where $f(\bX_{t-1}; \mathcal{G}, \btheta_\mu)$ is defined as in (\ref{FTGCN}), and $\varepsilon_{i,t}$ is the error term with zero mean.
Note that model (\ref{mean_model}) reduces to the TGCN model in \cite{Feng2019TemporalRR}, when the factors and their features are absent.
To estimate model (\ref{mean_model}), we consider the penalized least squares (PLS) estimator of $\btheta_\mu$ given by
\begin{align}\label{ols_loss}
\begin{split}
    \widehat{\btheta}_\mu &= \argmin_{\btheta_\mu}  \frac{1}{T} \sum_{t = 1}^T \Bigg( \frac{1}{N} \sum_{i = 1}^N  [r_{i,t} - f_i(\bX_{t-1}; \mathcal{G}, \btheta_\mu)]^2\\
    &\quad + \frac{\lambda^*}{N^2} \sum_{i = 1}^N \sum_{j = 1}^N \max\left\{0, -[f_i(\bX_{t-1}; \mathcal{G}, \btheta_\mu) - f_j(\bX_{t-1}; \mathcal{G}, \btheta_\mu)] (r_{i,t} - r_{j,t})\right\} \Bigg)\\
    & \equiv  \argmin_{\btheta_\mu} \frac{1}{T} \sum_{t = 1}^T \ell_{\mu}(\br_t, \bX_{t-1}; \mathcal{G}, \btheta_{\mu}, \lambda^*),
\end{split}
\end{align}
where the penalty term is utilized to ensure that the orders of $r_{i,t}$ and $r_{j,t}$ do not deviate largely from those of their predicted values, and $\lambda^*$ is a positive hyperparameter. As for $\widehat{\btheta}_\tau$, we adopt the Adam algorithm to compute
$\widehat{\btheta}_\mu$; see Algorithm \ref{alg:adam_mean} for the details. Using $\widehat{\btheta}_\mu$, we then estimate the conditional mean $\mu_{i,t}$ by
\begin{align}\label{hat_bmu_t}
    \widehat{\mu}_{i,t} = f_i(\bX_{t-1}; \mathcal{G}, \widehat{\btheta}_\mu)\,\,\, \mbox{ for }i=1,...,N.
\end{align}
It is worthwhile mentioning that the order-preserving penalty in (\ref{ols_loss}) has been widely adopted in the literature to improve the learning efficiency for conditional mean prediction \citep{zheng2007regression,Richard2013,Feng2019TemporalRR}. However,
this penalized method is inappropriate for the conditional quantile estimation, since
$Q_{i,t}(\tau)$ does not tend to be larger than $Q_{j,t}(\tau)$ when $r_{i,t}$ is larger than $r_{j,t}$.

\begin{algorithm}[h!]
\caption{The training procedure of $\widehat{\btheta}_\mu$ by the Adam algorithm.}\label{alg:adam_mean}
\begin{algorithmic}[1]
\REQUIRE ~~\\
    The sample: $\{(\bm{r}_{t}, \bX_{t-1})\}$;\\
    The initial value of network parameters in the FTGCN-based mean model: $\btheta_{\mu}^{(0)}$;\\
    The factor-augmented hypergraph: $\mathcal{G}$;\\
    Hyperparameters: $\lambda^*$, learning rate $\gamma$;\\
\STATE $l = 0$;
\REPEAT
\STATE $(\br_l, \bX_{l-1})$ $\leftarrow$ draw a random data point from $\{(\br_t, \bX_{t-1})\}$ (A cross-sectional minibatch);
\STATE $\bm{g}^{(l)} \leftarrow \nabla_{\btheta_\mu} \left[\ell_{\mu}(\br_l, \bX_{l-1}; \mathcal{G}, \btheta^{(l)}_\mu, \lambda^*) \right]$ (Gradients of minibatch estimator);
\STATE $\btheta_\mu^{(l + 1)} \leftarrow$ update parameters using learning rate $\gamma$ and gradients $\bm{g}^{(l)}$ (Adam);
\STATE $l \leftarrow l+1$;
\UNTIL{convergence of parameters $\btheta_\mu^{(l + 1)}$};
\ENSURE ~~\\
    The value of $\btheta_\mu^{(l + 1)}$, which is taken as the PLS estimator $\widehat{\btheta}_\mu$.
\end{algorithmic}
\end{algorithm}

\subsection{The QCM Learning for Higher-order Conditional Moments}\label{sec:qcm}
Let $\widehat{\bQ}_{t}(\tau_1),...,\widehat{\bQ}_{t}(\tau_K)$ be the vectors of estimated conditional quantiles at $K$ different quantile levels $\tau_1,...,\tau_K$, where $\widehat{\bQ}_{t}(\tau_k)\equiv (\widehat{Q}_{1,t}(\tau_k),...,\widehat{Q}_{N,t}(\tau_k))'$ for $k=1,...,K$ are computed as in (\ref{hat_bQ_t}). Below, we elaborate on how to estimate
$h_{i,t}$, $s_{i,t}$, and $k_{i,t}$ by the QCM method in \cite{QCM} for the fixed values of $i$ and $t$, based on $\widehat{Q}_{i,t}(\tau_1),...,\widehat{Q}_{i,t}(\tau_K)$.

The QCM method is motivated by the Cornish-Fisher expansion \citep{Fisher1938148MA}, which shows the following fundamental linkage between
conditional quantiles and conditional moments:
\begin{align}\label{cf_expansion}
\begin{split}
    Q_{i,t}(\tau_k) &= \mu_{i,t} + z(\tau_k) \sqrt{h_{i,t}} +  \left[z(\tau_k)^{2}-1\right] \frac{\sqrt{h_{i,t}} s_{i,t}}{6}
    \\
    &\quad +  \left[z(\tau_k)^{3}-3 z(\tau_k)\right] \frac{\sqrt{h_{i,t}}(k_{i,t}-3)}{24} + \sqrt{h_{i,t}}\omega_{i,t}(\tau_k)
\end{split}
\end{align}
for $k=1,...,K$, where $z(\tau_k)$ is the $\tau_k$-th quantile of standard normal distribution, and $\omega_{i,t}(\tau_k)$ is the remainder of this expansion. Define
\begin{align*}
&\varepsilon_{i,t,k}^{\bullet}=\varepsilon_{i,t,k}^{\ast}+\varepsilon_{i,t,k}^{\circ} \mbox{ with }\varepsilon_{i,t,k}^{\ast}=\sqrt{h_{i,t}}\omega_{i,t}(\tau_k)\mbox{ and }\varepsilon_{i,t,k}^{\circ}=\widehat{Q}_{i,t}(\tau_k)-Q_{i,t}(\tau_k),\\
&\bZ_k=\big(z(\tau_k), z(\tau_k)^2-1, z(\tau_k)^3-3z(\tau_k)\big)',\\
&\bbeta_{i,t}\equiv(\beta_{i,t,1}, \beta_{i,t,2}, \beta_{i,t,3})'=\Big(\sqrt{h_{i,t}}, \frac{\sqrt{h_{i,t}}s_{i,t}}{6}, \frac{\sqrt{h_{i,t}}(k_{i,t}-3)}{24}\Big)',
\end{align*}
where $\widehat{Q}_{i,t}(\tau_k)$ is the estimator of $Q_{i,t}(\tau_k)$. Then, we can rewrite (\ref{cf_expansion}) as follows:
\begin{align}\label{cf_expansion_1}
    \widehat{Q}_{i,t}(\tau_k) &= \mu_{i,t} + \bZ_k'\bbeta_{i,t}+\varepsilon_{i,t,k}^{\bullet}\,\,\, \mbox{ for }k=1,...,K,
\end{align}
where $\varepsilon_{i,t,k}^{\bullet}$ is the gross error containing the expansion error $\varepsilon_{i,t,k}^{\ast}$ and the quantile estimation error $\varepsilon_{i,t,k}^{\circ}$. Clearly, the equation (\ref{cf_expansion_1}) is a linear regression model with the response variable $\widehat{Q}_{i,t}(\tau_k)$, explanatory variables $\bZ_k$, parameter vector $(\mu_{i,t},\bbeta_{i,t}')'$, and error term $\varepsilon_{i,t,k}^{\bullet}$. Since $\varepsilon_{i,t,k}^{\bullet}$ may not have zero mean for model identification, we add an additional
deterministic intercept term $\gamma_{i,t}$ into the equation (\ref{cf_expansion_1}) to form the following linear regression model:
\begin{align}\label{cf_expansion_2}
\begin{split}
 \widehat{Q}_{i,t}(\tau_k) &= (\mu_{i,t}+\gamma_{i,t}) + \bZ_k'\bbeta_{i,t}+\varepsilon_{i,t,k}\\
&\equiv \bar{\bZ}_{k}'\btheta_{i,t}+\varepsilon_{i,t,k}\,\,\, \mbox{ for }k=1,...,K,
\end{split}
\end{align}
where $\varepsilon_{i,t,k}=\varepsilon_{i,t,k}^{\bullet}-\gamma_{i,t}$, $\bar{\bZ}_k=(1,\bZ_k')'$, and $\btheta_{i,t}=(\beta_{i,t,0},\bbeta_{i,t}')'$ with $\beta_{i,t,0}=\mu_{i,t}+\gamma_{i,t}$.

Let $\bY_{i,t}$ be a $K\times 1$ vector with entries $\widehat{Q}_{i,t}(\tau_k)$, $\bar{\bZ}$ be a $K\times 4$ matrix with rows $\bar{\bZ}_k'$, and $\bvarepsilon_{i,t}$ be a $K\times 1$ vector with entries $\varepsilon_{i,t,k}$. Then, the ordinary least squares (OLS) estimator of
$\btheta_{i,t}$ in (\ref{cf_expansion_2}) is
\begin{align}\label{ols}
\widehat{\btheta}_{i,t}\equiv(\widehat{\beta}_{i,t,0},\widehat{\bbeta}_{i,t}')'=(\bar{\bZ}'\bar{\bZ})^{-1}\bar{\bZ}'\bY_{i,t},
\end{align}
where $\widehat{\bbeta}_{i,t}=(\widehat{\beta}_{i,t,1}, \widehat{\beta}_{i,t,2}, \widehat{\beta}_{i,t,3})'$.
\cite{QCM} show that $\widehat{\btheta}_{i,t}\longrightarrow\btheta_{i,t}$ in probability as $K\to\infty$ under the following two classical conditions in the regression literature:

\begin{cond}\label{cond_1}
$\bar{\bZ}'\bar{\bZ}$ is positive definite.
\end{cond}

\begin{cond}\label{cond_2}
$\bar{\bZ}'\bvarepsilon_{i,t}/K\longrightarrow\pmb{0}$ in probability as $K\to\infty$.
\end{cond}

\noindent Consequently, by the continuous mapping theorem, we have
\begin{equation}\label{qcm}
\widehat{h}_{i,t} \equiv \widehat{\beta}_{i,t,1}^{\,2}\longrightarrow h_{i,t},\,\,\,
\widehat{s}_{i,t} \equiv \frac{6\widehat{\beta}_{i,t,2}}{\widehat{\beta}_{i,t,1}}\longrightarrow s_{i,t},\,\,\,\mbox{ and }\,\,\,
\widehat{k}_{i,t} \equiv \frac{24\widehat{\beta}_{i,t,3}}{\widehat{\beta}_{i,t,1}}+3\longrightarrow k_{i,t}
\end{equation}
in probability as $K\to\infty$, where $\widehat{h}_{i,t}$, $\widehat{s}_{i,t}$, and $\widehat{k}_{i,t}$ are the QCMs of
$h_{i,t}$, $s_{i,t}$, and $k_{i,t}$, respectively.
In order to make sure that $\widehat{h}_{i,t}$, $\widehat{s}_{i,t}$, and $\widehat{k}_{i,t}$ are moments under certain distribution of $r_{i,t}$, two necessary constraints are required:
\begin{align*}
    \widehat{h}_{i,t} \geq 0\,\,\, \text{and}\,\,\, \widehat{k}_{i,t} \geq \widehat{s}_{i,t}^2 + 1.
\end{align*}
Clearly, the first constraint holds automatically, and the second constraint can be checked directly based on the values of
$\widehat{k}_{i,t}$ and $\widehat{s}_{i,t}$. If the second constraint does not hold, we can easily replace
$\widehat{\btheta}_{i,t}$ in (\ref{ols}) by a constrained least squares estimator, so that the resulting
$\widehat{k}_{i,t}$ and $\widehat{s}_{i,t}$ satisfy this constraint; see more detailed discussions in  \cite{QCM}.
Moreover, it should be noted that we are unable to estimate $\mu_{i,t}$ by the QCM method. The reason is that $\mu_{i,t}$ can not be estimated by $\widehat{\beta}_{i,t,0}$ in (\ref{ols}) due to
the presence of $\gamma_{i,t}$. Therefore, we have to estimate $\mu_{i,t}$ separately by other methods (see, e.g., the graph-based method in Section \ref{sec_mean} above).

As we observed, the core idea of QCM method is to transform the estimation of conditional moments to that of conditional quantiles, giving us two remarkable advantages particularly in the realm of high-dimensional data analysis.

First, the QCM method is easy-to-implement, since it only requires the estimated conditional quantiles as the input to compute the OLS estimator $\widehat{\btheta}_{i,t}$. When $N$ is large, a direct estimation for the higher-order conditional moments $h_{i,t}$, $s_{i,t}$, and $k_{i,t}$ via high-dimensional GARCH-type models is computationally infeasible. The reason is that the high-dimensional GARCH-type models are fitted by the QML estimation method, which relies on a certain distribution of $\br_t$ to write down the log-likelihood function.
However, the log-likelihood function is too complex to be optimized for large $N$ cases. For example,
the often used Gaussian log-likelihood function depends on the inverse of many $N\times N$-dimensional variance-covariance matrices,
making its optimization infeasible.
Transforming the estimation of higher-order moments to that of quantiles circumvents this annoying difficulty, since the estimation of quantiles is a classical supervised learning but that of higher-order moments is not.
To be more specific, the supervised learning is a machine learning paradigm, and it aims to learn a function $f_0$ that maps features (say, $x$) to labels (say, $y$) supervised by a certain loss function without assuming the distribution of $x$ or $y$.
For example, the $\tau$-th quantile of $y$ can be learned by $f_0(x)$ using the quantile loss function $\rho_{\tau}(y-f_0(x))$ as the supervisor.
However, it is unclear how to design an appropriate loss function as the supervisor for learning the variance, skewness, or kurtosis of $y$ by $f_0(x)$, unless certain distributional assumption is made for $x$ or $y$.
This indicates that the direct estimation of $h_{i,t}$, $s_{i,t}$, and $k_{i,t}$ has to rely on a certain distribution of $r_{i,t}$, as done by the QML estimation in the high-dimensional GARCH-type models.
Owing to the supervised learning feature of quantiles, our indirect estimation of $h_{i,t}$, $s_{i,t}$, and $k_{i,t}$ from the QCM method does not need any distributional assumption of $r_{i,t}$, so it bypasses the computational difficulty raised in the direct estimation method to deal with large $N$ cases.

Second, the QCM method can largely reduce the risk of model mis-specification, since $\widehat{h}_{i,t}$, $\widehat{s}_{i,t}$, and $\widehat{k}_{i,t}$ are simultaneously computed without any prior estimation of $\mu_{i,t}$, and their consistency holds even when the specification of $Q_{i,t}(\tau)$ is mis-specified. This advantage is far beyond our expectations, since normally we have to first estimate $\mu_{i,t}$ and then $h_{i,t}$, $s_{i,t}$, and $k_{i,t}$ using some parametric models that are needed to be correctly specified to generate consistent estimators.
The reason leading to this advantage is that the QCM method is regression-based, so that
the impact of $\mu_{i,t}$ is eliminated by absorbing it into the intercept term and
the consistency of $\widehat{h}_{i,t}$, $\widehat{s}_{i,t}$, and $\widehat{k}_{i,t}$
is ensured by Conditions \ref{cond_1}--\ref{cond_2}. Note that Condition \ref{cond_1} holds for the often choices of quantile sequence $\{\tau_1,...,\tau_K\}$, and Condition \ref{cond_2} allows each model error $\varepsilon_{i,t,k}$ (including the quantile estimation error $\varepsilon_{i,t,k}^{\circ}$) to deviate from zero to some extent, as long as the averages of $\{\varepsilon_{i,t,k}\}$, $\{\varepsilon_{i,t,k}z(\tau_k)\}$, $\{\varepsilon_{i,t,k}[z(\tau_k)^2-1]\}$, and
$\{\varepsilon_{i,t,k}[z(\tau_k)^3-3z(\tau_k)]\}$ across $k$ are close to zero; see \cite{QCM} for more discussions on this aspect.

\subsection{Implementation Details of the GRACE Method}\label{sec:imple}

Due to the use of FTGCN, the GRACE method first needs to alleviate the risk of overfitting, a prevalent deficiency of the neural network.
Following the standard approach to circumvent overfitting, we chronologically partition the full data into three disjoint parts: training sample, validation sample, and testing sample. The training and validation samples are taken to do parameter estimation, and the testing sample is used to evaluate the truly out-of-sample performance of the GRACE method. To be more specific, we compute $\btheta_{\tau}^{(l)}$ at $l$-th iteration in Algorithm \ref{alg:adam_quantile} based on the training sample, and then calculate its corresponding validation sample error. Here,
the validation sample error is the value of the objective function in \eqref{qr_loss} based on the validation sample and
$\btheta_{\tau}=\btheta_{\tau}^{(l)}$. To regularize against overfitting, we utilize
the early stopping method to terminate the iteration process early in Algorithm \ref{alg:adam_quantile} when the validation sample error increases for several iterations, and select the estimator $\btheta_{\tau}^{(l)}$ having the smallest validation sample error as the quantile estimator $\widehat{\btheta}_\tau$. Similarly, the PLS estimator $\widehat{\btheta}_\mu$ is computed from the training and validation samples under
Algorithm \ref{alg:adam_mean}.

Next, the GRACE method uses $\widehat{\btheta}_{\tau_k}$ and $\widehat{\btheta}_\mu$ to predict the values of
$\widehat{Q}_{i,t}(\tau_k)$ and $\mu_{i,t}$ on the testing sample, respectively, where
 $\tau_k = k/(K+1)$, $k=1,...,K$, for simplicity.
In the large pool of $\widehat{Q}_{i,t}(\tau_k)$, some of $\widehat{Q}_{i,t}(\tau_k)$ are inevitable to be invalid.
Intuitively, it is reasonable to exclude those invalid $\widehat{Q}_{i,t}(\tau_k)$ for the computation of QCMs.
To achieve this goal, we make use of the unconditional coverage test $LR_{uc}$ in \cite{kupiec1995techniques} and conditional coverage test $LR_{cc}$ in \cite{Christoffersen1998EvaluatingIF}. Specifically, we compute $\widehat{Q}_{i,t}(\tau_k)$ on the training and validation samples, and apply $LR_{uc}$ and $LR_{cc}$ to detect whether the sequence of estimated conditional quantiles
 $\mathcal{Q}_{i}(\tau_k)\equiv\{\widehat{Q}_{i,t}(\tau_k): t\in \text{training and validation samples}\}$ is valid at the significance level $\alpha$ for each
 stock $i$ and quantile level $\tau_k$.
Then, we build a valid quantile level set for stock $i$:
\begin{equation}\label{omega_set}
\Omega_i = \{\tau_k: \text{ the validity of }\mathcal{Q}_{i}(\tau_k) \text{ is accepted by both } LR_{uc} \text { and } LR_{cc} \text{ at the level } \alpha\}.
\end{equation}
That is, $\Omega_i$ groups all of those quantile levels $\tau_k$, for which the sequence $\mathcal{Q}_{i}(\tau_k)$ is valid. Clearly, $\Omega_i$  depends on $\alpha$ and $K$ jointly, where its size (denoted by $|\Omega_i|$) is decreasing with the value of $\alpha$ while increasing with the value of $K$. In particular, we know that $|\Omega_i|=K$ when $\alpha=0$.
After having $\Omega_i$, we use the predicted values $\{\widehat{Q}_{i,t}(\tau_k): \tau_k\in\Omega_i\}$
to predict the values of $h_{i,t}$, $s_{i,t}$, and $k_{i,t}$ for stock $i$ via the related QCMs on the testing sample.
As $K$ is essentially replaced by $|\Omega_i|$ from the above manipulation, we need a large value of $|\Omega_i|$ to
ensure the consistency of the QCMs. This motivates us to discard those stocks having the value of $|\Omega_i|$ less than a predetermined tolerance $K_0$ (say, e.g., $K_0=30$).


Finally, the GRACE method employs different performance measures from the predicted values of $\mu_{i,t}$, $h_{i,t}$, $s_{i,t}$, and $k_{i,t}$
to construct portfolios, based on all of remaining stocks.

\section{Empirical Analysis}\label{sec:empirical}

\subsection{Data} \label{sec:data}
\begin{figure}
    \centering
    \includegraphics[width = \linewidth]{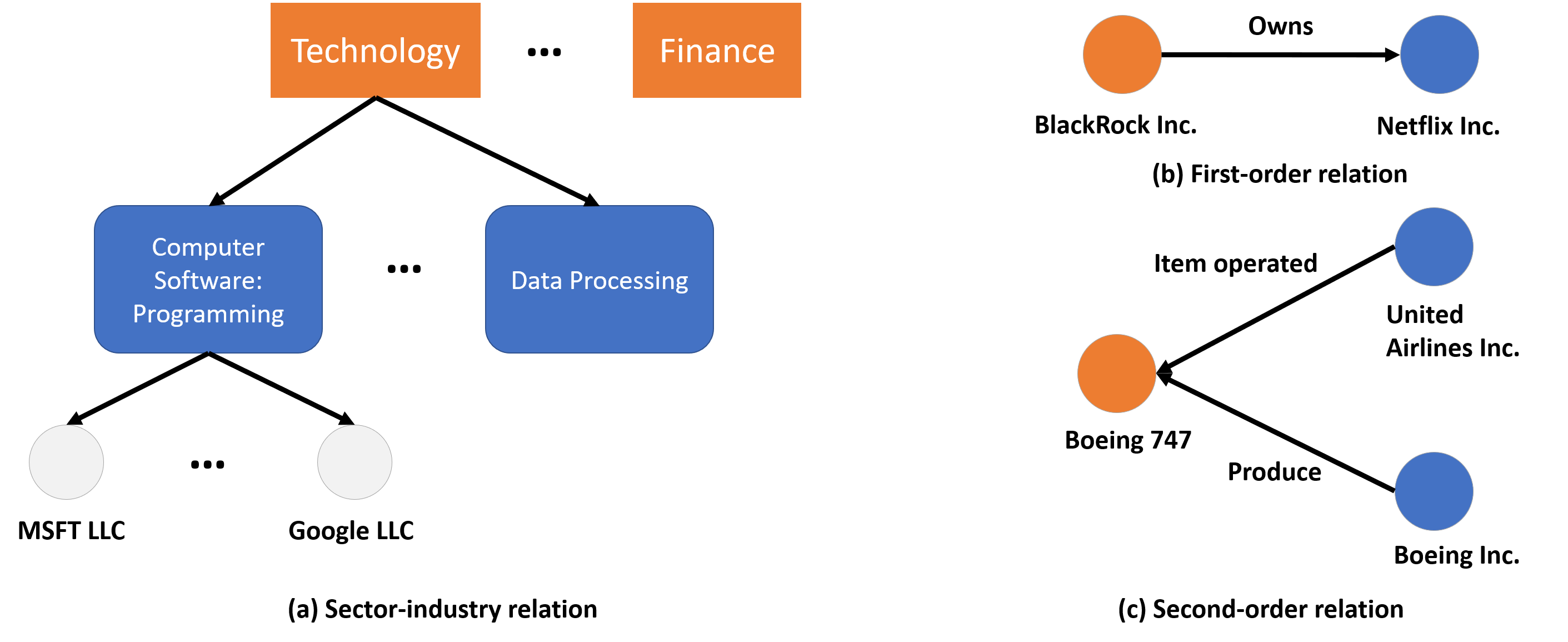}
    \caption{Examples of the sector-industry, first-order and second-order relations.}
    \label{fig:relation_data}
\end{figure}

We apply our GRACE method to construct portfolios based on the stocks in two major exchanges: NASDAQ and NYSE.
The stock data we consider are the same as those in \cite{Feng2019TemporalRR}, and they
contain daily prices from January 2, 2013 to December 8, 2017 for 1026 and 1737 stocks in NASDAQ and NYSE, respectively.
Alone with the stock price data, we also take the S2S relation data in \cite{Feng2019TemporalRR} to describe the multiple types of S2S relation. Based on the domain knowledge, the S2S relations can be divided into two groups: Sector-industry relations $\bm{E}_{stock}^{si}$ and Wiki company-based relations $\bm{E}_{stock}^{wiki}$ (see the Appendix A of \cite{Feng2019TemporalRR} for their detailed definitions).
Specifically, two stocks (say, stock $i$ and stock $j$) have a sector-industry relation if they belong to the same industry, where the industries are classified by the GICS standard. For example, all 1026 stocks in NASDAQ are divided into 13 different sectors, where each sector contains several industries; see Fig\,\ref{fig:relation_data}(a) for the sector-industry hierarchy of all 1026 stocks in this market. From this figure, we know that MSFT LLC and Google LLC  have an S2S relation since they belong to the same industry ``Computer Software: Programming''.

%
%
%

Meanwhile, two stocks can also have a Wiki company-based relation if they have an either first-order or second-order relation. The first-order and second-order relations have the format of ``$\mbox{company }i\overset{R}{\longrightarrow}\mbox{company }j$'' and ``$\mbox{company }i\overset{R_1}{\longrightarrow}\mbox{entity }k\overset{R_2}{\longleftarrow}\mbox{company }j$'', respectively, where
the companies $i$ and $j$ bridged by an entity $k$ are corresponding to the stocks $i$ and $j$, respectively,  and
the relations $R$, $R_1$, and $R_2$ are defined in Wikidata (\textit{https://www.wikidata.org/wiki/Wikidata:List\_of\_properties/all}). It turns out that there are 5 and 53 different types of first-order and second-order relations, respectively.
Fig\,\ref{fig:relation_data}(b) and \ref{fig:relation_data}(c) give some illustrating examples on the first-order and second-order relations. We see from this figure that BlackRock Inc. has a first-order relation with Netflix Inc. since BlackRock Inc. owns Netflix  Inc., and United Airlines Inc.
and Boeing Inc. have a second-order relation since Boeing Inc. produces Boeing 747 that is sold to United Airlines Inc. In sum, Table \ref{table:relation} lists the number of S2S relation types and the ratio of S2S relations to all possible stock pairs in NASDAQ and NYSE. Since the ratio of S2S relations is always less than 10\%, it indicates that the S2S relations in both markets are sparse.

\begin{table}[h]
\begin{center}
\begin{minipage}{\textwidth}
\caption{Summary of S2S relations}\label{table:relation}
\begin{tabular*}{\textwidth}{@{\extracolsep{\fill}}ccccc@{\extracolsep{\fill}}}
\toprule%
& \multicolumn{2}{@{}c@{}}{Sector-industry relation} & \multicolumn{2}{@{}c@{}}{Wiki company-based relation} \\
\cmidrule{2-3}\cmidrule{4-5}%
Market &Relation types & Relation ratio (pairwise) &Relation types & Relation ratio (pairwise)\\
\midrule
NASDAQ &112 &5.00$\%$ &42 &$0.21\%$\\
NYSE &130 &9.37$\%$ &32 &$0.30\%$\\
\bottomrule
\end{tabular*}
\end{minipage}
\end{center}
\end{table}

In addition, the F2S relation data are also needed to facilitate our GRACE method.
Among a great variety of factors, we use the prevalent daily Fama-French five factors \citep{fama2015five} to specify the related F2S relations $\bm{E}_{factor}^{ff5}$, based on the asset-pricing knowledge. These five factors are excess market return, RMW, HML, SMB, and CMA, and their daily data can be downloaded from the homepage of Kenneth French (\textit{https://mba.tuck.dartmouth.edu/pages/faculty/ken.french/data\_library.html}).


For stock $i$, we let $r_{i,t}$ be its 1-day return and $r_{i,t}^{(k_*)}=\frac{1}{k_*}\sum_{s=1}^{k_*}r_{i,t+1-s}$ be its
$k_*$-day moving average of returns at day $t$, and use the OLS method to calculate its $5$-dimensional vector of factor exposures
(denoted by $\lambda_{i,t}=(\lambda_{i,t}^{(1)},...,\lambda_{i,t}^{(5)})'$) based on the sample of 1-day returns in a half-year rolling window up to day $t$.
For factor $b$, we similarly let
$f_{b,t}$ be its value, $f_{b,t}^{(k_*)}$ be its $k_*$-day moving average, and
$\bar{\lambda}_{b,t}^{(b_*)}$ be its exposure on factor $b_*$ at day $t$, where
we assume that factor $b$ has one exposure on itself and zero exposure on other factors (w.r.t., $b_*\not=b$).
Now, based on ten stock features $r_{i,t}^{(k_*)}$ and $\lambda_{i,t}^{(b_*)}$ together with ten factor features $f_{b,t}^{(k_*)}$ and $\bar{\lambda}_{b,t}^{(b_*)}$
for $k_*=1,5,10,20,30$ and $b_*=1,2,3,4,5$, our feature tensor $\bX_{t-1}=[\bX_{1,t-1},...,\bX_{N,t-1},\bX_{N+1,t-1},...,\bX_{N+B,t-1}]$ in (\ref{bQ_t}) is taken as
\begin{equation}\label{feature_input}
\bX_{i,t-1}=\begin{pmatrix}
r_{i,t-S}^{(1)} & \cdots & r_{i,t-1}^{(1)}\\
\cdots  & \cdots & \cdots \\
r_{i,t-S}^{(30)} & \cdots & r_{i,t-1}^{(30)} \\
\lambda_{i,t-S}^{(1)} & \cdots & \lambda_{i,t-1}^{(1)} \\
\cdots & \cdots & \cdots \\
\lambda_{i,t-S}^{(5)} & \cdots & \lambda_{i,t-1}^{(5)}
\end{pmatrix}
\mbox{ and }
\bX_{N+b,t-1}=\begin{pmatrix}
f_{b,t-S}^{(1)} & \cdots & f_{b,t-1}^{(1)}\\
\cdots & \cdots & \cdots \\
f_{b,t-S}^{(30)} & \cdots & f_{b,t-1}^{(30)} \\
\bar{\lambda}_{b,t-S}^{(1)} & \cdots & \bar{\lambda}_{b,t-1}^{(1)} \\
\cdots & \cdots & \cdots \\
\bar{\lambda}_{b,t-S}^{(5)} &  \cdots & \bar{\lambda}_{b,t-1}^{(5)}
\end{pmatrix},
\end{equation}
for $i=1,...,N$ and $b=1,...,B$, where $N=1026$ (or 1737) for the NASDAQ (or the NYSE) market, $B=5$, $\bX_{i,t-1}\in\mathcal{R}^{P\times S}$, and
$\bX_{N+b,t-1}\in\mathcal{R}^{P\times S}$ with $P=10$. It is worthy noting that
each entry in $\bX_{i,t-1}$ and $\bX_{N+b,t-1}$ is normalized by its range in the training sample
to reduce its skewness and leptokurtosis; see the similar implementation in
\cite{Feng2019TemporalRR}.

With the full data sample $\{\br_{t},\bX_{t-1}\}$ in hand, we divide it into three disjoint parts in the same way as \cite{Feng2019TemporalRR}: The training sample from January 2, 2013 to December 31, 2015 has 756 trading days, the validation sample follows and ends on December 30, 2016 with 252 trading days, and the testing sample covers the remaining  237 trading days from January 3, 2017 to December 8, 2017 (i.e., the out-of-sample period). Now, based on the values of
hyperparameters listed in Table \ref{tab:hyperparameters}, our GRACE method is implemented as the details specified in Section \ref{sec:imple} above.
Here, the selection of $\lambda^*$, $\gamma$, $S$, and $d$ for the FTGCN-based models is suggested by the tuning results in
\cite{Feng2019TemporalRR}, and that of $\alpha$ and $K$ for the QCM method will be examined in the sequel.


\begin{table}[h]
\begin{center}
\begin{minipage}{\textwidth}
\caption{List of hyperparameters.}
\begin{tabular*}{\textwidth}{@{\extracolsep{\fill}}llr@{\extracolsep{\fill}}}
\toprule%
Hyperparameter & Description    & Value   \\
    \hline
    $\lambda^*$   & tuning parameter in the computation of PLS estimator  & $0.1$ \\
    $\gamma$    & learning rate in the Adam algorithm & $10^{-3}$\\
    $S$  & length of lagged features in the LSTM & $16$ \\
    $d$  & number of hidden units in the LSTM &  $64$ \\
    $\alpha$ & significance level of $LR_{uc}$ and $LR_{cc}$ tests & 0.01 \\
    $K$         & number of predetermined quantile levels   &$199$ \\
\bottomrule
\end{tabular*}\label{tab:hyperparameters}
\end{minipage}
\end{center}
\end{table}



\subsection{Comparison Methods}
Besides our GRACE method, other graph-based methods can also be adopted to select portfolios using the same idea, except for different
models to predict the conditional quantiles and mean of $r_{i,t}$. Below, we introduce two alternative graph-based methods for the purpose of comparison.

The first competitor is labeled as GRACE$_{1}$, which
replaces the factor-augmented hypergraph in the GRACE method with the hypergraph in \cite{Feng2019TemporalRR}, and leaves other
mechanisms (including the input features and the selection of tuning hyperparameters)
unchanged. The comparison between the GRACE and GRACE$_1$ methods is to verify whether incorporating the asset pricing knowledge into the hypergraph is informative for portfolio selection.

The second competitor is the simple GRACE (denoted by GRACE$_2$) method, which is motivated by the network autoregression model in \cite{zhu2017network} and the FDNQR model in \cite{xu2022dynamic}. Specifically,
the GRACE$_2$ method predicts the conditional quantiles of $r_{i,t}$ based on the following specification:
\begin{align}\label{quantile_xu}
    Q_{i,t}(\tau) = \alpha_{\tau} + \gamma_{\tau} \sum_{j=1}^{N} w_{i,j} r_{j,t-1} + \bm{\zeta}'_{\tau} \bx_{i, t-1} + \bvarsigma_{1,\tau}'\vF_{t-1},
\end{align}
where $\bx_{i, t-1}$ is the last column of $\bX_{i,t-1}$ in (\ref{feature_input}), $\vF_{t-1}$ is the 5-dimensional vector containing the values of Fama-French five factors at $t-1$, and other notations are inherited from (\ref{NQAR}). By construction, model (\ref{quantile_xu}) uses the term $\bm{\zeta}'_{\tau} \bx_{i, t-1}$ (replacing the term $\bbeta_{\tau} \bz_i + \zeta r_{i,t-1}$ in (\ref{NQAR})) to account for the lag-$1$ stock features, and it takes the term $\bvarsigma_{1,\tau}'\vF_{t-1}$ to
include the lag-$1$ factor features.
Similarly, the GRACE$_2$ method predicts the conditional mean of $r_{i,t}$ by the following factor-augmented network autoregressive specification:
\begin{align}\label{mean_xu}
    r_{i,t} = \alpha_{\mu} + \gamma_{\mu} \sum_{j=1}^{N} w_{i,j} r_{j,t-1} + \bm{\zeta}_{\mu}' \bx_{i, t-1} + \bvarsigma_{1,\mu}'\vF_{t-1} + \varepsilon_{i,t}^*,
\end{align}
where $\varepsilon_{i,t}^*$ is the error term with zero mean, and $\alpha_{\mu}$, $\gamma_{\mu}$, $\bm{\zeta}_{\mu}$, and $\bvarsigma_{1,\mu}$ are unknown regression coefficients.
Here, $w_{i,j}=a_{i,j}/n_i$ in models (\ref{quantile_xu}) and (\ref{mean_xu}) is determined by $\bm{E}_{stock}$ in the GRACE method, such that
$a_{i,j}=1$ if there is any S2S relation between stock $i$ and stock $j$, and $a_{i,j}=0$ otherwise.
We estimate models (\ref{quantile_xu}) and (\ref{mean_xu}) respectively via the the quantile loss and $L_2$ loss functions using the data from the combination of training and validation samples, and then proceed the portfolio selection on the testing sample.
One may extend both models to contain the stock and factor features up to lag-$S$. However, our unreported analysis shows that this extension
makes model estimation less stable, leading to the worse performance in portfolio selection.
Clearly, the comparison between the GRACE and GRACE$_2$ methods aims to exhibit how inadequate are those simple settings in models (\ref{quantile_xu}) and (\ref{mean_xu}) for portfolio selection.

\subsection{Economic Performance Evaluation}

This subsection evaluates the out-of-sample performance of the long-short portfolios selected by the GRACE method.
After sorting all stocks via a certain performance measure into 10 deciles, the long-short portfolio is re-balanced on every trading day
via buying the $10\%$ highest ranking stocks (decile 10) and selling the $10\%$ lowest ranking stocks (decile 1) with equal weights.
The performance measures to sort all stocks include the M (mean), MV (Mean--variance), MVSK (Mean--variance with skewness and kurtosis), SR (Sharpe ratio), and SRSK (Sharpe ratio with skewness and kurtosis), where the definitions of the last four performance measures are given in (\ref{eqn_1})--(\ref{eqn_4}), and
the values of all five performance measures are computed based on the predicted values of $\mu_{i,t}$, $h_{i,t}$, $s_{i,t}$, and $k_{i,t}$ from the GRACE method with $\bm{E}_{stock}=\bm{E}_{stock}^{wiki}$, $\bm{E}_{factor}=\bm{E}_{factor}^{ff5}$, and hyperparameters
taken as in Table \ref{tab:hyperparameters}.
For the performance measures MV, MVSK, and SRSK, the hyperparameters $\lambda_1$, $\lambda_2$, and $\lambda_3$ are tuned by
the grid search within  the sets $A_1, A_2$, and $A_3$, respectively, to maximize the value of SR of each long-short portfolio re-balanced on training and validation samples. Here, due to different ranges of $h_{i,t}$ and $s_{i,t}$ (or $k_{i,t}$), we take
\begin{align*}
A_1&=\{a\times 10^{-b}: a=1,2,...,9\mbox{ and }b=-1,0,...,3\},\\
A_2&=A_3=\{a\times 10^{-b}: a=1,2,...,9\mbox{ and }b=2,3,...,6\}.
\end{align*}
As a comparison, the GRACE$_1$ and GRACE$_2$ methods are also used to select the long-short portfolios under the similar procedure as above, where
the GRACE$_1$ method with the performance measure M is the benchmark method proposed by \cite{Feng2019TemporalRR}.
Since the QCM method used by each method relies on the choices of $\alpha$ and $K$, the stock pool for portfolio selection under the performance measures MV, MVSK, SR, and SRSK varies with the choice of $\alpha$ or $K$.
For the sake of consistency, the portfolio selection under the performance measure $M$ will use the same stock pool
as for other performance measures in the sequel, although the implementation of performance measure $M$ is independent of higher-order conditional moments and not affected by the choice of $\alpha$ or $K$ technically.


From an economic viewpoint, we compare all of the selected portfolios in terms of their annualized SR,
which is the ratio of annualized excess return to annualized risk.
To compute the annualized excess return, we use the Treasury bill rate as a proxy for the risk-free return, and
take the transaction cost of $3\text{\textperthousand}$ (i.e., 30 basic points)  for buying and selling into account as done in \cite{engle2012}.
Note that when there is no ambiguity, we omit the wording ``annualized'' below for ease of presentation.


\subsubsection{Impacts of Method and Performance Measure}\label{sec:pm}

We first assess how the graph-based methods and performance measures affect the out-of-sample performance of portfolios. Table \ref{tab:NYSE} reports the values of (excess) return, risk, and SR of the out-of-sample portfolios
selected by three different methods with five different performance measures in NASDAQ and NYSE.
From this table, we can have the following findings:

\begin{table}[!h]
\begin{center}
\begin{minipage}{\textwidth}
\caption{Out-of-sample performances of long-short portfolios across different methods and performance measures.}\label{tab:NYSE}
\begin{tabular*}{\textwidth}{@{\extracolsep{\fill}}llccccccc@{\extracolsep{\fill}}}  
    \toprule
                   &       & \multicolumn{3}{c}{NASDAQ} &       & \multicolumn{3}{c}{NYSE} \\
\cmidrule{3-5}\cmidrule{7-9}    Method & Measure    & Return ($\%$) & Risk ($\%$)  & SR    &       & Return ($\%$) & Risk ($\%$)  & SR \\
\midrule
    GRACE & M     & 25.40 & 8.53  & 2.98  &       & 29.04 & 8.74  & 3.10 \\
          & MV    & 38.81 & 10.43 & 3.72  &       & 29.36 & 8.62  & 3.17 \\
          & MVSK  & 37.04 & 8.43  & 4.39  &       & 30.01 & 8.62  & 3.25 \\
          & SR    & 37.48 & 9.03  & 4.15  &       & 13.90 & 3.43  & 3.47 \\
          & SRSK  & 39.21 & 8.15  & 4.81  &       & 13.95 & 3.43  & 3.48 \\
\cmidrule{2-9}
    GRACE$_1$ & M     & 12.08 & 8.45  & 1.43  &       & 25.74  & 8.97  & 2.87  \\
          & MV    & 13.55 & 7.83  & 1.73  &       & 25.82  & 8.97  & 2.88  \\
          & MVSK  & 33.45 & 8.24  & 4.06  &       & 25.82  & 8.97  & 2.88  \\
          & SR    & 26.57 & 6.48  & 4.10  &       & 16.91  & 5.10  & 3.32  \\
          & SRSK  & 42.09 & 8.86  & 4.75  &       & 16.97  & 5.10  & 3.33  \\
\cmidrule{2-9}
    GRACE$_2$ & M     & 0.06  & 8.59  & 0.01  &       & -2.09 & 8.94  & -0.46 \\
          & MV    & 4.55  & 10.57 & 0.43  &       & 0.45  & 8.80  & -0.18 \\
          & MVSK  & 6.66  & 8.50  & 0.78  &       & 2.26  & 8.79  & 0.03 \\
          & SR    & 4.53  & 9.11  & 0.50  &       & 3.00  & 3.51  & 0.28 \\
          & SRSK  & 5.90  & 8.19  & 0.72  &       & 3.56  & 3.51  & 0.45 \\
        \bottomrule
\end{tabular*}
\end{minipage}
\end{center}
\end{table}

\begin{itemize}
\item[(i)] For the GRACE method, the SRSK and M portfolios have the largest and smallest values of SR, respectively, in both markets, implying the necessity of using three higher-order conditional moments for portfolio selection.
    Particularly, the advantage of using higher-order conditional moments is more evident in NASDAQ than NYSE by observing that the value of SR for the SRSK portfolio is $61\%$ (or $12\%$) higher than that for the M portfolio in NASDAQ (or NYSE). Moreover, the values of return and risk indicate that the SRSK portfolio has a larger value of SR mainly because it can generate a much larger (or smaller) value of return (or risk) than the M portfolio in NASDAQ (or NYSE). Another difference between NASDAQ and NYSE is the influence of conditional skewness and kurtosis.
    Specifically, using the conditional skewness and kurtosis can well decrease the portfolio risk in NASDAQ while only marginally increase the portfolio return in NYSE, according to the comparison between MV and MVSK (or SR and SRSK) portfolios. The aforementioned distinction in two markets may attribute to the fact that the NYSE has a relatively more ``normal'' environment than the NASDAQ, so that the function of higher-order conditional moments (particularly the conditional skewness and kurtosis) is relatively weaker for portfolio selection.

\item[(ii)] For the GRACE$_1$ and GRACE$_2$ methods, both of them perform worse than the GRACE method, regardless of the choice of performance measure. The advantage of GRACE method over GRACE$_1$ method is exceptionally significant for M and MV portfolios in NASDAQ, since
    the M and MV portfolios selected by GRACE method have 108\% and 115\% higher value of SR than those selected by GRACE$_1$ method, respectively. This finding shows that incorporating asset price information is
    more important for portfolio selection in NASDAQ than NYSE, especially when the conditional skewness and kurtosis are not taken into account. Moreover, the value of SR for the best portfolio selected by the GRACE method is 236\% and 21\% higher than that selected by the benchmark method (i.e., the GRACE$_1$ method with the performance measure M) in NASDAQ and NYSE, respectively.
    In all cases, the GRACE$_2$ method has a much worse performance than other two methods. This is not unexpected, because the simple model settings in the GRACE$_2$ method can not capture the effects of features on stock returns adequately.
\end{itemize}

Overall, the above findings clearly demonstrate the importance of higher-order conditional moments
as well as the asset pricing knowledge in portfolio selection through the GRACE method.

\subsubsection{Impact of $\pmb{E}_{stock}$}\label{sec:rel_imp}
Since all considered three graph-based methods depend on the S2S relation set $\bm{E}_{stock}$, a natural question is what kind of
S2S relation set is more informative for portfolio selection. To answer this question, we alter all three methods by choosing
 $\bm{E}_{stock}=\bm{E}_{stock}^{si}$ or $\bm{E}_{stock}^{all}$ while keeping other settings as for Table \ref{tab:NYSE} unchanged, where $\bm{E}_{stock}^{all}$ is the union set of $\bm{E}_{stock}^{wiki}$ and $\bm{E}_{stock}^{si}$.
Table \ref{tab:SR_relation} reports the values of out-of-sample SR for portfolios selected from three different choices of $\bm{E}_{stock}$.
From Table \ref{tab:SR_relation}, we find that the value of SR for the $\bm{E}_{stock}^{si}$-based portfolio
is smaller than that for the corresponding $\bm{E}_{stock}^{wiki}$-based portfolio in all cases, except for the M portfolios selected by the GRACE$_1$ method in NASDAQ. Particularly, the advantage of $\bm{E}_{stock}^{wiki}$-based portfolio over $\bm{E}_{stock}^{si}$-based portfolio is much more substantial for the GRACE$_1$ method with the performance measures MVSK, SR, and SRSK in NASDAQ.
This finding indicates that the S2S relations in $\bm{E}_{stock}^{si}$
could be less informative than those in $\bm{E}_{stock}^{wiki}$ to learn higher-order conditional moments, especially
when the asset pricing knowledge is absent.

Moreover, we find from Table \ref{tab:SR_relation} that using a richer S2S relation set $\bm{E}_{stock}^{all}$ to replace the single S2S relation set $\bm{E}_{stock}^{wiki}$ gives no change or little change to the values of SR in the GRACE and GRACE$_1$ methods, and this replacement even makes the portfolios have smaller values of SR for many cases in the GRACE$_1$ method. The reason is probably that the long-term correlations between stocks are largely driven by the factors through the F2S relations in  $\bm{E}_{factor}^{ff5}$, and they could be wrongly captured by the S2S relations in $\bm{E}_{stock}^{si}$ when the asset pricing knowledge is absent. For example, the stocks belonging to the sector ``Basic Industries'' tend to have large market capitalization, while the SMB factor in \cite{fama2015five} represents the outperformance of small-cap stocks over large-cap ones during a long-term. Hence, the comovement of stocks in the sector ``Basic Industries'' is more properly captured by the F2S relations with respect to the SMB factor in $\bm{E}_{factor}^{ff5}$ rather than the sector-industry relations in $\bm{E}_{stock}^{si}$.
The unsatisfactory performance from the use of $\bm{E}_{stock}^{all}$ becomes more evident in the
 GRACE$_2$ method. This conveys the information that it is inappropriate to
ignore the type of S2S relation as done by models (\ref{quantile_xu})--(\ref{mean_xu}), when the domain knowledge on multiple types of S2S relation is available.

In sum, we could reach a general conclusion that $\bm{E}_{stock}^{si}$ is less informative than $\bm{E}_{stock}^{wiki}$ for portfolio selection.
Hence, if $\bm{E}_{stock}^{wiki}$ is accessible, we recommend it for practical use.


\begin{table}[!h]
\begin{minipage}{\textwidth}
\caption{Out-of-sample SRs of long-short portfolios across different choices of $\bm{E}_{stock}$.}\label{tab:SR_relation}%
\begin{tabular*}{\textwidth}{@{\extracolsep{\fill}}llccccccc@{\extracolsep{\fill}}}
    \toprule
          &       & \multicolumn{3}{c}{NASDAQ} &       & \multicolumn{3}{c}{NYSE} \\
\cmidrule{3-5}\cmidrule{7-9}
Method & Measure    & $\bm{E}_{stock}^{si}$   & $\bm{E}_{stock}^{wiki}$  & $\bm{E}_{stock}^{all}$   &       & $\bm{E}_{stock}^{si}$   & $\bm{E}_{stock}^{wiki}$  & $\bm{E}_{stock}^{all}$  \\
\cmidrule{1-9}
GRACE & M     & 2.74  & 2.98  & 2.98  &       & 3.04  & 3.10  & 3.12 \\
          & MV    & 2.86  & 3.72  & 3.74  &       & 3.04  & 3.17  & 3.17 \\
          & MVSK  & 3.98  & 4.39  & 4.40  &       & 3.04  & 3.25  & 3.25 \\
          & SR    & 3.81  & 4.15  & 4.18  &       & 3.04  & 3.47  & 3.47 \\
          & SRSK  & 4.52  & 4.81  & 4.81  &       & 3.13  & 3.48  & 3.48 \\
\cmidrule{2-9}
GRACE$_1$ & M     & 1.58  & 1.43  & 1.52  &       & 2.76  & 2.88  & 2.87 \\
          & MV    & 1.60  & 1.73  & 1.55  &       & 2.77  & 2.88  & 2.87 \\
          & MVSK  & 2.94  & 4.06  & 4.02  &       & 2.77  & 2.88  & 2.88 \\
          & SR    & 2.14  & 4.10  & 4.01  &       & 3.19  & 3.32  & 3.32 \\
          & SRSK  & 3.19  & 4.75  & 4.65  &       & 3.19  & 3.32  & 3.32 \\
\cmidrule{2-9}
GRACE$_2$ & M     & -0.17 & 0.01  & -1.24 &       & -1.88 & -0.46 & -2.81 \\
          & MV    & 0.12  & 0.43  & -1.02 &       & -1.77 & -0.18 & -2.80 \\
          & MVSK  & 0.66  & 0.78  & -0.55 &       & -1.71 & 0.03  & -2.78 \\
          & SR    & 0.43  & 0.50  & -1.34 &       & -1.56 & 0.28  & -2.63 \\
          & SRSK  & 0.61  & 0.72  & -1.29 &       & -1.33 & 0.45  & -2.58 \\
    \bottomrule
\end{tabular*}%
\end{minipage}
\end{table}%

\subsubsection{Impact of $\alpha$ and $K$}

Before implementing the QCM method, a valid quantile level set $\Omega_i$ in (\ref{omega_set}) is defined for each stock $i$.
As discussed in Section \ref{sec:imple} above, we only use the predicted conditional quantiles at those quantile levels in $\Omega_i$ to predict higher-order conditional moments.
Intuitively, when the value of $K$ is fixed (i.e., the quantile sequence $\{\tau_1,...,\tau_K\}$ is given),
a large value of $\alpha$ enhances the reliability of the conditional quantile prediction, but at the same time, it reduces the learning efficiency of the higher-order conditional moments as the size of $\Omega_i$ becomes small. Hence, there is a trade-off between reliability and learning efficiency, in terms of the choice of $\alpha$.  To address this issue, we examine the impact of $\alpha$ on the performance of portfolio by
changing the value of $\alpha$ while keeping other settings in the same way as for Table \ref{tab:NYSE}.

Specifically, we select the long-short portfolios for each method when $\alpha \in \{0\%, 1\%, 5\%, 10\%\}$, and report the related results of
 out-of-sample SR in Table  \ref{table:level}.
From this table, we find that
except for the M and SR portfolios from the GRACE$_1$ method in NASDAQ, all of other portfolios achieve the maximum value of SR at $\alpha=1\%$. When the value of $\alpha$ decreases from $1\%$ to $0\%$ (i.e., $|\Omega_i|=K$),
the performance of most portfolios becomes worse, particularly for those selected by the GRACE$_1$ method in NYSE.
This finding indicates that the tests $LR_{uc}$ and $LR_{cc}$ can effectively exclude those extremely ill-behaved predictions of conditional quantiles to improve the performance of QCM method. When the value of $\alpha$ increases from $1\%$ to $5\%$ or $10\%$, the performance of most portfolios also becomes worse but with different tendencies in two markets. For example, the reduction in SR under GRACE$_1$ method is much larger than that under the GRACE method in NASDAQ, whereas the reduction in SR has similar patterns in NYSE based on both methods. This observation shows the advantage of the factor-augmented hypergraph, which stabilizes the performance of the GRACE method across the choice of $\alpha$.

 Besides $\alpha$, $K$ is also related to the construction of $\Omega_i$ for all three methods. A large value of $K$ is needed to ensure the consistency of the QCMs in theory, however, it is inevitable to increase the cost in computation as
the conditional quantile model has to be trained $K$ different times.
Clearly, the choice of $K$ reflects the trade-off between the computational cost and learning efficiency.
To study the impact of $K$, we select the long-short portfolios as for Table \ref{tab:NYSE} but with different values of $K$, and report the values of out-of-sample SR for those portfolios in Table \ref{tab:tau_num}.
The findings from this table generally match our expectation.
First, the value of SR increases with the value of $K$ for each portfolio. This is because a larger value of $K$
can grab more information on the conditional distribution of $r_{i,t}$ by learning its conditional quantiles at more refined quantile levels.
Second, the performance of portfolios from the GRACE method is more stable across the choice of $K$ than that from the GRACE$_1$ method. This
finding is consistent to that in Table \ref{table:level}.
Third, the values of SR under the GRACE$_2$ method are always less than one or even below zero in many cases, implying the inadequacy of
models (\ref{quantile_xu})--(\ref{mean_xu}) for learning conditional quantiles and mean of $r_{i,t}$.

Overall, the portfolios from the GRACE method show a robust performance
over the choice of $\alpha$ and $K$, and they outperform those from the GRACE$_1$ and GRACE$_2$ methods
for all examined choices of $\alpha$ and $K$.
In practice, we recommend to take $\alpha=1\%$ and $K=199$ for the GRACE method, since this seems a desirable choice
to balance the reliability (or computational cost) and learning efficiency, as demonstrated by the aforementioned results.

\begin{table}[h]
\begin{center}
\begin{minipage}{\textwidth}
\caption{Out-of-sample SRs of long-short portfolios across different choices of $\alpha$.}\label{table:level}
\begin{tabular*}{\textwidth}{@{\extracolsep{\fill}}llccccccccc@{\extracolsep{\fill}}}
    \toprule
          &       & \multicolumn{4}{c}{NASDAQ}    &       & \multicolumn{4}{c}{NYSE} \\
\cmidrule{3-6}\cmidrule{8-11}
Method & Measure    & 0\%   & 1\%   & 5\%   & 10\%  &       & 0\%   & 1\%   & 5\%   & 10\% \\
\midrule
GRACE  & M     & 2.35  & 2.98  & 2.83  & 2.81  &       & 2.88  & 3.10  & 3.06  & 3.05 \\
          & MV    & 3.58  & 3.72  & 3.67  & 3.61  &       & 3.01  & 3.17  & 3.11  & 3.09 \\
          & MVSK  & 4.17  & 4.39  & 4.21  & 4.09  &       & 3.03  & 3.25  & 3.22  & 3.21 \\
          & SR    & 3.99  & 4.15  & 3.94  & 3.87  &       & 3.11  & 3.47  & 3.45  & 3.42 \\
          & SRSK  & 4.78  & 4.81  & 4.76  & 4.62  &       & 3.15  & 3.48  & 3.43  & 3.43 \\
\cmidrule{2-11}
GRACE$_1$  & M     & 1.42  & 1.43  & 1.34  & 1.46  &       & 1.46  & 2.88  & 2.81  & 2.83 \\
          & MV    & 1.48  & 1.73  & 1.62  & 1.59  &       & 1.53  & 2.88  & 2.81  & 2.83 \\
          & MVSK  & 3.77  & 4.06  & 4.03  & 2.98  &       & 1.67  & 2.88  & 2.81  & 2.83 \\
          & SR    & 4.61  & 4.10  & 3.81  & 2.11  &       & 2.11  & 3.32  & 3.24  & 3.22 \\
          & SRSK  & 4.73  & 4.75  & 4.68  & 3.06  &       & 2.76  & 3.32  & 3.25  & 3.26 \\
\cmidrule{2-11}
GRACE$_2$  & M     & 0.01  & 0.01  & -0.02 & -0.04 &       & -0.63 & -0.46 & -0.49 & -0.48 \\
          & MV    & 0.22  & 0.43  & 0.31  & 0.26  &       & -0.41 & -0.18 & -0.22 & -0.26 \\
          & MVSK  & 0.68  & 0.78  & 0.67  & 0.65  &       & -0.14 & 0.03  & -0.08 & -0.12 \\
          & SR    & 0.31  & 0.50  & 0.41  & 0.38  &       & 0.17  & 0.28  & 0.21  & 0.18 \\
          & SRSK  & 0.64  & 0.72  & 0.59  & 0.53  &       & 0.25  & 0.45  & 0.31  & 0.27 \\
    \bottomrule
\end{tabular*}
\end{minipage}
\end{center}
\end{table}

\begin{table}[h]
\begin{center}
\begin{minipage}{\textwidth}
\caption{Out-of-sample SRs of long-short portfolios across different choices of $K$.}
\begin{tabular*}{\textwidth}{@{\extracolsep{\fill}}llccccccc@{\extracolsep{\fill}}}
\toprule
&       & \multicolumn{3}{c}{NASDAQ} &       & \multicolumn{3}{c}{NYSE} \\
\cmidrule{3-5}\cmidrule{7-9}    Method & Measure & 49    & 99    & 199   &       & 49    & 99    & 199 \\
\midrule
GRACE & M     & 2.88  & 2.91  & 2.98  &       & 3.07  & 3.08  & 3.10 \\
          & MV    & 3.16  & 3.51  & 3.72  &       & 3.07  & 3.12  & 3.17 \\
          & MVSK  & 3.22  & 3.64  & 4.39  &       & 3.19  & 3.21  & 3.25 \\
          & SR    & 3.59  & 3.77  & 4.15  &       & 3.37  & 3.42  & 3.47 \\
          & SRSK  & 3.63  & 3.92  & 4.81  &       & 3.43  & 3.44  & 3.48 \\
\cmidrule{2-9}
GRACE$_1$ & M     & 1.01  & 1.31  & 1.43  &       & 1.73  & 2.44  & 2.88 \\
          & MV    & 1.13  & 1.52  & 1.73  &       & 1.78  & 2.51  & 2.88 \\
          & MVSK  & 3.18  & 3.44  & 4.06  &       & 2.61  & 2.71  & 2.88 \\
          & SR    & 3.09  & 3.59  & 4.10  &       & 2.43  & 3.04  & 3.32 \\
          & SRSK  & 3.47  & 3.76  & 4.75  &       & 2.87  & 3.04  & 3.32 \\
\cmidrule{2-9}
GRACE$_2$ & M     & -0.12 & -0.03 & 0.01  &       & -0.50 & -0.49 & -0.46 \\
          & MV    & 0.30  & 0.38  & 0.43  &       & -0.24 & -0.20 & -0.18 \\
          & MVSK  & 0.47  & 0.61  & 0.78  &       & -0.22 & -0.19 & 0.03 \\
          & SR    & 0.29  & 0.42  & 0.50  &       & -0.08 & 0.05  & 0.28 \\
          & SRSK  & 0.49  & 0.62  & 0.72  &       & -0.02 & 0.09  & 0.45 \\
    \bottomrule
\end{tabular*}\label{tab:tau_num}
\end{minipage}
\end{center}
\end{table}


\subsubsection{Impact of Transaction Cost}

The transaction cost is a non-negligible important factor for portfolio selection.
In NASDAQ and NYSE, the cost of a transaction mainly consists of three components: Commission, stamp tax, and slippage.
While the rates of commission and stamp tax are easily accessible, the slippage is hard to be measured quantitatively
as the liquidity issue could make it difficult to execute transactions at a pre-specified price without affecting the market price, especially for small-cap stocks. Normally, the transaction cost of 30 basis points is appropriate to capture the total effect of
commission, stamp tax, and slippage. However, the slippage could be larger in some extreme circumstances. Therefore, it is necessary to conservatively consider some higher transaction costs than 30 basis points to investigate their impact on portfolio selection.

To achieve this goal, Table \ref{table:cost} reports the values of out-of-sample SR for long-short portfolios selected from each method, when
the transaction costs are $30$, $50$, $75$, and $100$ basis points.
From this table, we can obtain some interesting findings.
First, as expected, the value of SR for each portfolio decreases with the transaction cost.
Second, regardless of transaction cost, the SRSK portfolio is the best one followed successively by the SR, MVSK, MV, and M portfolios
in the GRACE and GRACE$_1$ methods.
Moreover, the SRSK portfolio from the GRACE method has a more stable performance to the level of transaction cost than other portfolios in NASDAQ.
Third, the GRACE method outperforms the GRACE$_1$ method in all considered cases, except that the SR and SRSK portfolios from the GRACE$_1$ method
perform better than those from the GRACE method when the transaction cost is 100 basis points in NYSE.
Fourth, the GRACE$_2$ method performs much worse than the GRACE method, and its performance becomes more unsatisfactory
when the transaction cost is higher.

Overall, we find that the advantage of using the higher-order conditional moments together with the asset pricing knowledge
from the GRACE method for portfolio selection is unchanged with the setting of transaction cost.

\begin{table}[!h]
\begin{center}
\begin{minipage}{\textwidth}
\caption{Out-of-sample SRs of long-short portfolios across different transaction costs (in basis points).}\label{table:cost}
\begin{tabular*}{\textwidth}{@{\extracolsep{\fill}}llccccccccc@{\extracolsep{\fill}}}
\toprule%
          &       & \multicolumn{4}{c}{NASDAQ}    &       & \multicolumn{4}{c}{NYSE} \\
\cmidrule{3-6}\cmidrule{8-11}    Method & Measure & 30    & 50    & 75    & 100   &       & 30    & 50    & 75    & 100 \\
    \midrule
    GRACE   & M     & 2.98  & 1.48  & 0.54  & -0.40 &       & 3.10  & 1.79  & 1.25  & 0.83 \\
          & MV    & 3.72  & 2.88  & 1.97  & 0.71  &       & 3.17  & 1.80  & 1.27  & 0.83 \\
          & MVSK  & 4.39  & 3.12  & 2.04  & 0.87  &       & 3.25  & 1.84  & 1.27  & 0.85 \\
          & SR    & 4.15  & 3.46  & 2.69  & 1.95  &       & 3.47  & 2.45  & 1.38  & 0.90 \\
          & SRSK  & 4.81  & 4.01  & 3.13  & 2.30  &       & 3.48  & 2.51  & 1.39  & 0.92 \\
\cmidrule{2-11}
GRACE$_1$  & M     & 1.43  & 1.14  & 0.29  & -0.77 &       & 2.88  & 1.44  & 1.01  & 0.45 \\
          & MV    & 1.73  & 1.60  & 0.78  & 0.44  &       & 2.88  & 1.50  & 1.16  & 0.46 \\
          & MVSK  & 4.06  & 2.71  & 1.63  & 0.65  &       & 2.88  & 1.52  & 1.19  & 0.48 \\
          & SR    & 4.10  & 3.11  & 2.48  & 1.23  &       & 3.32  & 2.11  & 1.22  & 0.92 \\
          & SRSK  & 4.75  & 3.82  & 2.53  & 1.49  &       & 3.32  & 2.33  & 1.24  & 0.94 \\
\cmidrule{2-11}
GRACE$_2$  & M     & 0.01  & -1.28 & -2.33 & -2.89 &       & -0.46 & -0.78 & -1.33 & -1.87 \\
          & MV    & 0.43  & -0.27 & -1.85 & -2.19 &       & -0.18 & -0.73 & -0.98 & -1.47 \\
          & MVSK  & 0.78  & -0.21 & -1.71 & -1.97 &       & 0.03  & -0.68 & -0.90 & -1.32 \\
          & SR    & 0.50  & 0.04  & -0.36 & -0.75 &       & 0.28  & -0.40 & -0.74 & -1.29 \\
          & SRSK  & 0.72  & 0.26  & -0.31 & -0.48 &       & 0.45  & -0.37 & -0.72 & -1.22 \\
    \bottomrule
\end{tabular*}
\end{minipage}
\end{center}
\end{table}

\subsection{Statistical Performance Evaluation}

From an economic viewpoint, the good performance of our GRACE method in portfolio selection has been demonstrated above. From a statistical viewpoint, it is worthwhile to check whether the GRACE method can well estimate and predict the conditional moments of $r_{i,t}$, shedding light on
its success in portfolio selection.

\subsubsection{The Comovement of Conditional Moments}

\begin{figure}[!h]
    \centering
    \includegraphics[width = \linewidth, height=0.45\textheight]{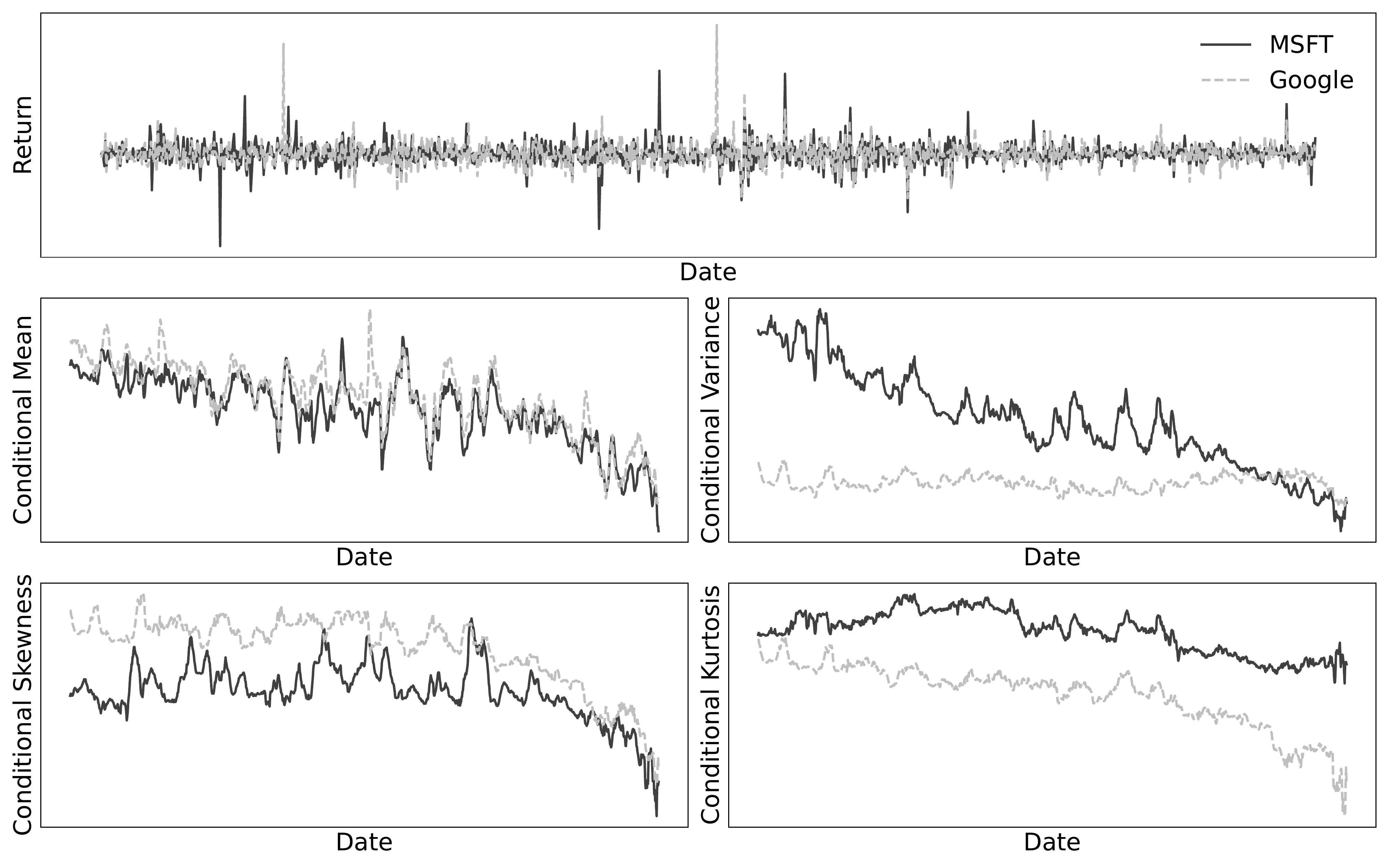}
    \caption{The plots of $r_{i,t}$, $\widehat{\mu}_{i,t}$, $\widehat{h}_{i,t}$, $\widehat{s}_{i,t}$ and $\widehat{k}_{i,t}$ for MSFT LLC and Google LLC from January 2, 2013 to December 8, 2017.}
    \label{fig:conditional}
\end{figure}

Intuitively, the conditional moments of two linked stocks should have a tendency of comovement. Since the number of
stocks is large, there has no easy way to exhibit whether most pairs of linked stocks have this tendency. For ease of
illustration, we only plot $r_{i,t}$, $\widehat{\mu}_{i,t}$, $\widehat{h}_{i,t}$, $\widehat{s}_{i,t}$, and $\widehat{k}_{i,t}$ of MSFT LLC and Google LLC based on the GRACE method in Fig\,\ref{fig:conditional}.
Here, $\widehat{\mu}_{i,t}$, $\widehat{h}_{i,t}$, $\widehat{s}_{i,t}$, and $\widehat{k}_{i,t}$ are estimated values during the in-sample period (i.e., the period of the training and validation samples), and they are predicted values during the out-of-sample period.
 From Fig\,\ref{fig:conditional}, we find that not only the stock returns but also all estimated and predicted conditional moments have similar trends during the entire period.

\subsubsection{The Validity of Conditional Moments}

Since the true values of conditional moments are unobserved, no explicit quantity can measure the precision of  $\widehat{h}_{i,t}$, $\widehat{s}_{i,t}$, and $\widehat{k}_{i,t}$.
To circumvent this deficiency, we propose some hypothesis tests to detect whether $\widehat{\mu}_{i,t}$, $\widehat{h}_{i,t}$, $\widehat{s}_{i,t}$, and $\widehat{k}_{i,t}$ are valid by using a similar idea as in
 \cite{Gu2020EmpiricalAP}. Specifically, we define $\alpha_{i}^{\mu} = E(e_{i,t}^{\mu})$, $\alpha_{i}^{h} = E(e_{i,t}^{h})$, $\alpha_{i}^{s} = E(e_{i,t}^{s})$, and $\alpha_{i}^{k} = E(e_{i,t}^{k})$, where $e_{i,t}^\mu = r_{i,t} - \mu_{i,t}$, $e_{i,t}^h = (r_{i,t} - \mu_{i,t})^2 - h_{i,t}$,
 $e_{i,t}^s = \left (\frac{r_{i,t} - \mu_{i,t}}{\sqrt{h_{i,t}}}\right )^3 - s_{i,t}$, and $e_{i,t}^k = \left(\frac{r_{i,t} - \mu_{i,t}}{\sqrt{h_{i,t}}}\right )^4 - k_{i,t}$
for $i=1,...,N$. Based on the estimates $\{\widehat{e}_{i,t}^{\mu}\}_{t=1}^{T}$ with $\widehat{e}_{i,t}^{\mu}=r_{i,t}-\widehat{\mu}_{i,t}$, we  adopt the classical Student's $t$ test $\mathbb{T}_{i}^{\mu}$ to
detect the null hypothesis $\mathbb{H}_{i}^{\mu}: \alpha_{i}^{\mu} = 0$.
If $\mathbb{H}_{i}^{\mu}$ is not rejected by $\mathbb{T}_{i}^{\mu}$ at the significance level $\alpha^*$, then we regard that $\{\widehat{\mu}_{i,t}\}_{t=1}^{T}$ is valid.
Similarly, the Student's $t$ tests $\mathbb{T}_{i}^{h}$, $\mathbb{T}_{i}^{s}$, and $\mathbb{T}_{i}^{k}$
to detect the null hypotheses $\mathbb{H}_{i}^{h}: \alpha_{i}^{h} = 0$, $\mathbb{H}_{i}^{s}: \alpha_{i}^{s} = 0$, and $\mathbb{H}_{i}^{k}: \alpha_{i}^{k} = 0$, respectively, can be used to examine the validity of $\{\widehat{\mu}_{i,t}\}_{t=1}^{T}$, $\{\widehat{h}_{i,t}\}_{t=1}^{T}$, $\{\widehat{s}_{i,t}\}_{t=1}^{T}$, and $\{\widehat{k}_{i,t}\}_{t=1}^{T}$.

We apply each Student's $t$ test to check the validity of conditional moments of stock $i$ at the significance level $\alpha^*\in\{1\%, 5\%, 10\%\}$, and then report the percentage of stocks having valid conditional moments in Table \ref{tab:t_test}. From this table,
we first find that the percentage of stocks having valid conditional moments during the out-of-sample period is higher than that during
the in-sample period in most cases. This may indicate that none of the methods has the problem of overfitting.
Next, except for the out-of-sample results of $\mathbb{T}_{i}^{s}$ in NYSE, all of the testing results show
that both GRACE and GRACE$_1$ methods deliver much better estimated and predicted conditional moments than the GRACE$_2$ method.
This statistically explains why both GRACE and GRACE$_1$ methods can select better portfolios than the GRACE$_2$ method.
Moreover, we observe that the GRACE method
performs better than GARCE$_1$ method according to the results at the significance level of 5\% and 10\%, although
both methods have the same results at the significance level of 1\%.
This advantage of GRACE method over GARCE$_1$ method shows the necessity of using factor-augmented hypergraph
for learning conditional moments.


\begin{table}[!h]
  \centering
  \caption{Percentages of stocks having valid conditional moments.}
  \setlength{\tabcolsep}{1.2mm}{
    \begin{tabular}{lrrrrrrrrrrrrrrrrrrrrrrr}
    \toprule
\cmidrule{2-24}          & \multicolumn{11}{c}{NASDAQ}
       &       & \multicolumn{11}{c}{NYSE} \\
\cmidrule{2-12}\cmidrule{14-24}          & \multicolumn{3}{c}{GRACE} &       & \multicolumn{3}{c}{GRACE$_1$} &       & \multicolumn{3}{c}{GRACE$_2$} &       & \multicolumn{3}{c}{GRACE} &       & \multicolumn{3}{c}{GRACE$_1$} &       & \multicolumn{3}{c}{GRACE$_2$} \\
\cmidrule{2-4}\cmidrule{6-8}\cmidrule{10-12}\cmidrule{14-16}\cmidrule{18-20}\cmidrule{22-24}    Test  & 1\%   & 5\%   & 10\%  &       & 1\%   & 5\%   & 10\%  &       & 1\%   & 5\%   & 10\%  &       & 1\%   & 5\%   & 10\%  &       & 1\%   & 5\%   & 10\%  &       & 1\%   & 5\%   & 10\% \\
\cmidrule{1-24}

          & \multicolumn{23}{c}{Panel A: In-sample period} \\

$\mathbb{T}_{i}^{\mu}$    & 95.0  & 71.8  & 57.6  &       & 95.0  & 71.7  & 56.6  &       & 87.2  & 65.3  & 49.0  &       & 96.4  & 83.4  & 75.4  &       & 96.4  & 82.8  & 71.9  &       & 78.1  & 75.9  & 70.0  \\
    $\mathbb{T}_{i}^{h}$     & 79.4  & 72.0  & 69.0  &       & 79.4  & 71.4  & 67.0  &       & 71.8  & 64.7  & 61.0  &       & 82.9  & 77.5  & 74.3  &       & 82.9  & 76.8  & 73.2  &       & 73.8  & 67.9  & 65.5  \\
    $\mathbb{T}_{i}^{s}$     & 86.7  & 82.1  & 73.4  &       & 86.7  & 80.6  & 71.5  &       & 77.2  & 73.3  & 68.0  &       & 86.6  & 82.7  & 76.8  &       & 86.6  & 81.4  & 75.1  &       & 75.6  & 63.6  & 61.2  \\
    $\mathbb{T}_{i}^{k}$     & 90.3  & 73.9  & 61.2  &       & 90.3  & 73.6  & 59.6  &       & 84.2  & 72.3  & 60.5  &       & 94.7  & 83.2  & 76.6  &       & 94.7  & 82.6  & 71.0  &       & 72.2  & 71.3  & 65.8  \\
& \multicolumn{23}{c}{Panel B: Out-of-sample period} \\
$\mathbb{T}_{i}^{\mu}$    & 94.0  & 85.7  & 78.3  &       & 94.0  & 83.4  & 74.9  &       & 79.3  & 77.1  & 65.8  &       & 93.2  & 80.9  & 73.9  &       & 93.2  & 78.8  & 66.8  &       & 86.8  & 71.3  & 50.8  \\
    $\mathbb{T}_{i}^{h}$     & 94.1  & 86.4  & 82.4  &       & 94.1  & 86.4  & 80.8  &       & 82.6  & 78.7  & 72.1  &       & 93.4  & 89.4  & 86.2  &       & 93.4  & 87.3  & 82.8  &       & 88.5  & 81.4  & 66.4  \\
    $\mathbb{T}_{i}^{s}$     & 89.6  & 82.3  & 74.8  &       & 89.6  & 82.0  & 73.2  &       & 92.8  & 78.2  & 76.5  &       & 71.5  & 66.8  & 61.4  &       & 71.5  & 63.7  & 56.1  &       & 85.6  & 78.9  & 72.6  \\
    $\mathbb{T}_{i}^{k}$     & 99.3  & 94.9  & 86.1  &       & 99.3  & 94.9  & 84.5  &       & 98.5  & 92.1  & 71.1  &       & 99.0  & 94.6  & 85.9  &       & 99.0  & 94.6  & 85.9  &       & 71.3  & 72.1  & 62.4  \\
    \bottomrule
    \end{tabular}%
    }
  \label{tab:t_test}%
\end{table}%

\section{Concluding Remarks}\label{sec:conclusion}

This paper proposes a new GRACE method for big portfolio selection under different performance measures that
are defined by four conditional moments (with respect to mean, variance, skewness, and kurtosis) of stock returns.
The GRACE method builds on the FTGCN and the QCM method: The former embeds the factor-augmented hypergraph within a graph neural network to obtain the estimates of mean and quantiles, and the latter transforms the estimates of quantiles into those of higher-order moments.

The most attractive feature of the GRACE method is its capacity to estimate conditional variance, skewness, and kurtosis of  high-dimensional stock returns, so the big portfolios under four performance measures MV, MVSK, SR, and SRSK can be constructed from thousands of stocks or even more. There are two reasons leading to this feature. First, the FTGCN takes the domain knowledge on S2S relations and the asset pricing knowledge on F2S relations to form the factor-augmented hypergraph. The knowledge of S2S and F2S relations is sparse to make estimation tractable in high-dimension, and at the same time, it is informative to capture the interplay between two stocks as well as the driving force from common factors in all stocks. Second, the QCM method only needs the estimates of quantiles to obtain those of higher-order moments, and the quantiles can be estimated by a supervised learning without assuming any distribution of stock returns.
If we intend to estimate variance, skewness, and kurtosis directly, a specific distribution of stock returns
is inevitably needed, as done by the traditional method based on high-dimensional GARCH models.
However, due to the presence of distribution of stock returns, the estimation of high-dimensional GARCH models becomes computationally infeasible when the dimension of stocks is large.

The importance of our GRACE method is further demonstrated by the empirical studies in NASDAQ and NYSE.
From an economic viewpoint, we find that the SRSK portfolio selected by the GRACE method earns the out-of-sample SR of $4.81$ and $3.48$ in NASDAQ and NYSE, respectively,
and it beats the M portfolio (with the out-of-sample SR of $1.43$ in NASDAQ and $2.87$ in NYSE) selected by the benchmark method in \cite{Feng2019TemporalRR} by a wide margin. Moreover, we find that regardless of the performance measure, the portfolios from the GRACE method have more stable and larger values of out-of-sample SR than those from the competing methods, across different settings of hyperparameter and transaction cost. From a statistical viewpoint, we find that the predicted conditional moments from the GRACE method are largely valid and capable of reflecting the comovement between linked stocks.

In future, our GRACE method can be extended in several directions. First, it is interesting to design some new hypergraphs to incorporate
subjective information from the experiences and beliefs of investors) or unstructured data information from financial news and social media contents \citep{KKX,FXZ}. Second, it is intriguing to examine whether other observed factors or data-driven factors \citep{giglio2022factor}
are useful for specifying the F2S relations. Third, it is desirable to
utilize the information of S2S and F2S relations for proposing GARCH-type methods \citep{Engle2019LargeDC,pakel2021fitting}, which
are computationally feasible to learn
higher-order moments when the dimension of the data is very large.

\bibliographystyle{imsart-nameyear}
\bibliography{Ref}

\end{document}